%% file: main.tex
\documentclass{article}

\usepackage{arxiv}

\usepackage[utf8]{inputenc} 
\usepackage[T1]{fontenc}    
\usepackage{url}            
\usepackage{booktabs}       
\usepackage{amsfonts}       
\usepackage{nicefrac}       
\usepackage{microtype}      
\usepackage{lipsum}
\usepackage{graphicx}

\usepackage{amsmath}
\usepackage{amsthm}
\usepackage{amssymb}
\usepackage{mathtools}
\usepackage{arydshln}
\RequirePackage{algorithm}
\RequirePackage{algorithmic}

\usepackage[accsupp]{axessibility}  

\usepackage[hidelinks]{hyperref}

\input{includes/base}
\input{includes/commands}

\input{includes/symbols}

\begin{document}

\title{EPTQ: Enhanced Post-Training Quantization via Hessian-guided Network-wise Optimization} 

\author{
 Ofir Gordon \\
  Sony Semiconductor Israel\\
  \texttt{ofir.gordon@sony.com} \\
   \And
 Elad Cohen \\
  Sony Semiconductor Israel\\
  \texttt{elad.cohen@sony.com}
  \And
 Hai Victor Habi \\
  Sony Semiconductor Israel\\
  \texttt{hai.habi@sony.com}
\And
 Arnon Netzer \\
  Sony Semiconductor Israel\\
  \texttt{arnon.netzer@sony.com}
}

\input{files/open}
\input{files/introduction}

\input{files/background}
\input{files/preliminaries}

\input{files/method}
\input{files/experimental}

\input{files/conclusions}
\input{files/ref}
\input{files/appendix}

\end{document}

%% file: includes/base.tex
\usepackage{microtype}
\usepackage{booktabs} 

\usepackage{color-edits} 
\addauthor[Hai]{hvh}{blue}
\addauthor[Ofir]{ofir}{red}

\newtheorem{assumption}{Assumption}
\newtheorem{proposition}{Proposition}

\usepackage{svg}
\svgsetup{inkscapeformat=png}
\usepackage{bm}
\usepackage{float}
\usepackage{subcaption}
\usepackage{multirow}

%% file: includes/commands.tex
\newcommand\MakeUppercaseGreek[1]{
  \begingroup
    \let\psi\Psi
    \let\omega\Omega
    \let\gamma\Gamma
    \MakeUppercase{#1}
  \endgroup}
\newcommand{\vectorsym}[1]{\bm{#1}}
\newcommand{\expectation}[2]{\mathbb{E}_{#2}\left[#1\right]}

\newcommand{\normaldis}[2]{\mathcal{N}(#1,#2)}
\newcommand{\brackets}[1]{\left(#1\right)}
\newcommand{\abs}[1]{\left|#1\right|}
\newcommand{\norm}[1]{\left\lVert#1\right\rVert}

\newcommand{\matsym}[1]{\mathbf{#1}}
\newcommand{\squareb}[1]{\left[{#1}\right]}

\newcommand{\divc}[2]{\frac{\partial#1}{\partial#2}}

\newcommand{\at}[2]{\left.#1\right\vert_{#2}}

\newcommand{\tstd}[1]{{\color{gray}\tiny $\pm#1$}}

%% file: includes/symbols.tex
\newcommand{\name}{EPTQ}
\newcommand{\brecq}{BRECQ~\cite{li2021brecq}}
\newcommand{\adaround}{AdaRound~\cite{nagel2020up}}
\newcommand{\adaquant}{AdaQuant~\cite{hubara2021accurate}}

\newcommand{\qdrop}{{QDrop~\cite{wei2021qdrop}}}
\newcommand{\nwq}{{NWQ~\cite{zheng2022leveraging}}}
\newcommand{\pdquant}{{PD-Quant~\cite{liu2023pd}}}

\newcommand{\our}{{\name} (Ours)}

\newcommand{\numipts}{n}

\newcommand{\resdelta}{\Delta_{\downarrow}}

\newcommand{\kdm}{\mathcal{L}_{KD}}
\newcommand{\kdmInput}[1]{\kdm\brackets{#1}}

\newcommand{\z}[0]{\vectorsym{z}}
\newcommand{\y}[0]{\vectorsym{y}}
\newcommand{\x}[0]{\vectorsym{x}}
\newcommand{\w}[0]{\vectorsym{w}}

\newcommand{\lossdiff}{\Delta\mathcal{L}}
\newcommand{\hmse}[0]{\textsc{Hmse}}
\newcommand{\maxhess}[1]{\lambda^{\brackets{#1}}}
\newcommand{\hessian}[1]{\matsym{H}^{(\vectorsym{#1})}}
\newcommand{\squarejac}{\vectorsym{u}}

\newcommand{\layer}{\ell}
\newcommand{\chan}{j}
\newcommand{\wf}{\mathbf{\w}}
\newcommand{\lw}{\wf^{(\layer)}}
\newcommand{\idxw}[1]{\wf^{(#1)}}
\newcommand{\quantlw}{\tilde{\wf}^{(\layer)}}
\newcommand{\zf}{\mathbf{\z}}
\newcommand{\lz}{\zf^{(\layer)}}
\newcommand{\idxz}[1]{\zf^{(#1)}}
\newcommand{\quantlz}{\tilde{\zf}^{(\layer)}}

%% file: files/open.tex
\maketitle
\begin{abstract}
    Quantization is a key method for deploying deep neural networks on edge devices with limited memory and computation resources. Recent improvements in Post-Training Quantization (PTQ) methods were achieved by an additional local optimization process for learning the weight quantization rounding policy. 
    However, a gap exists when employing network-wise optimization with small representative datasets. In this paper, we propose a new method for enhanced PTQ (EPTQ) that employs a network-wise quantization optimization process,
    which benefits from considering cross-layer dependencies during optimization. EPTQ enables network-wise optimization with a small representative dataset using a novel sample-layer attention score based on a label-free Hessian matrix upper bound.  
    The label-free approach makes our method suitable for the PTQ scheme.
    We give a theoretical analysis for the said bound and use it to construct a knowledge distillation loss that guides the optimization to focus on the more sensitive layers and samples.
    In addition, we leverage the Hessian upper bound to improve the weight quantization parameters selection by focusing on the more sensitive elements in the weight tensors. 
    Empirically, by employing {\name} we achieve state-of-the-art results on various models, tasks, and datasets, including ImageNet classification, COCO object detection, and Pascal-VOC for semantic segmentation.
\end{abstract}

%% file: files/introduction.tex
\section{Introduction}

Deep Neural Network (DNN) models are used to solve a wide variety of real-world computer vision tasks, such as Image Classification~\cite{he2016resnet, howard2017mobilenets, sandler2018mobilenetv2, radosavovic2020regnet}, Object Detection~\cite{lin2017focal, wong2019yolo}, and Semantic Segmentation~\cite{chen2018deeplab}.  
Indeed, in recent years there has been a significant increase in the deployment of DNN models on edge devices.
However, deploying these networks remains challenging due to their considerable memory usage and computational requirements.

To tackle this challenge, various methods have been explored, including the design of hardware-dedicated architectures \cite{gholami2018squeezenext, cai2018proxylessnas, cai2019once}, pruning techniques \cite{dong2017learning, he2018amc, huang2018data, yu2022hessian}, and quantization approaches \cite{banner2018scalable, banner2018, kim2020position, gholami2021survey}.
In this work, we focus on quantization, a practical approach that effectively reduces the size of the model and its computational demands. This is achieved by converting the weight and activation tensors into low-bit-width representations. 

In general, there are two primary methods for quantizing a neural network model.
The first one is Post-Training Quantization (PTQ)~\cite{banner2019aciq, hubara2020improving, habi2021hptq, li2021brecq, yao2022rapq}, which performs a small statistics collection to determine the quantization parameters without the need for labeled data. 
PTQ is valued for its simplicity and efficient runtime 
but it may suffer from accuracy degradation. 
The second approach is Quantization-Aware Training (QAT)~\cite{gupta2015deep, jacob2018quantization, habi2020hmq, esser2020lsq, nagel2022overcoming}. 
In QAT, the model parameters undergo retraining with labeled data and the task loss function, to mitigate quantization-induced errors.
Although QAT typically achieves better accuracy than PTQ, it comes with an increased computational and algorithmic complexity because of the necessity for training over the entire dataset.

\begin{figure}[tb]
\centering
    \includegraphics[width=0.88\textwidth]{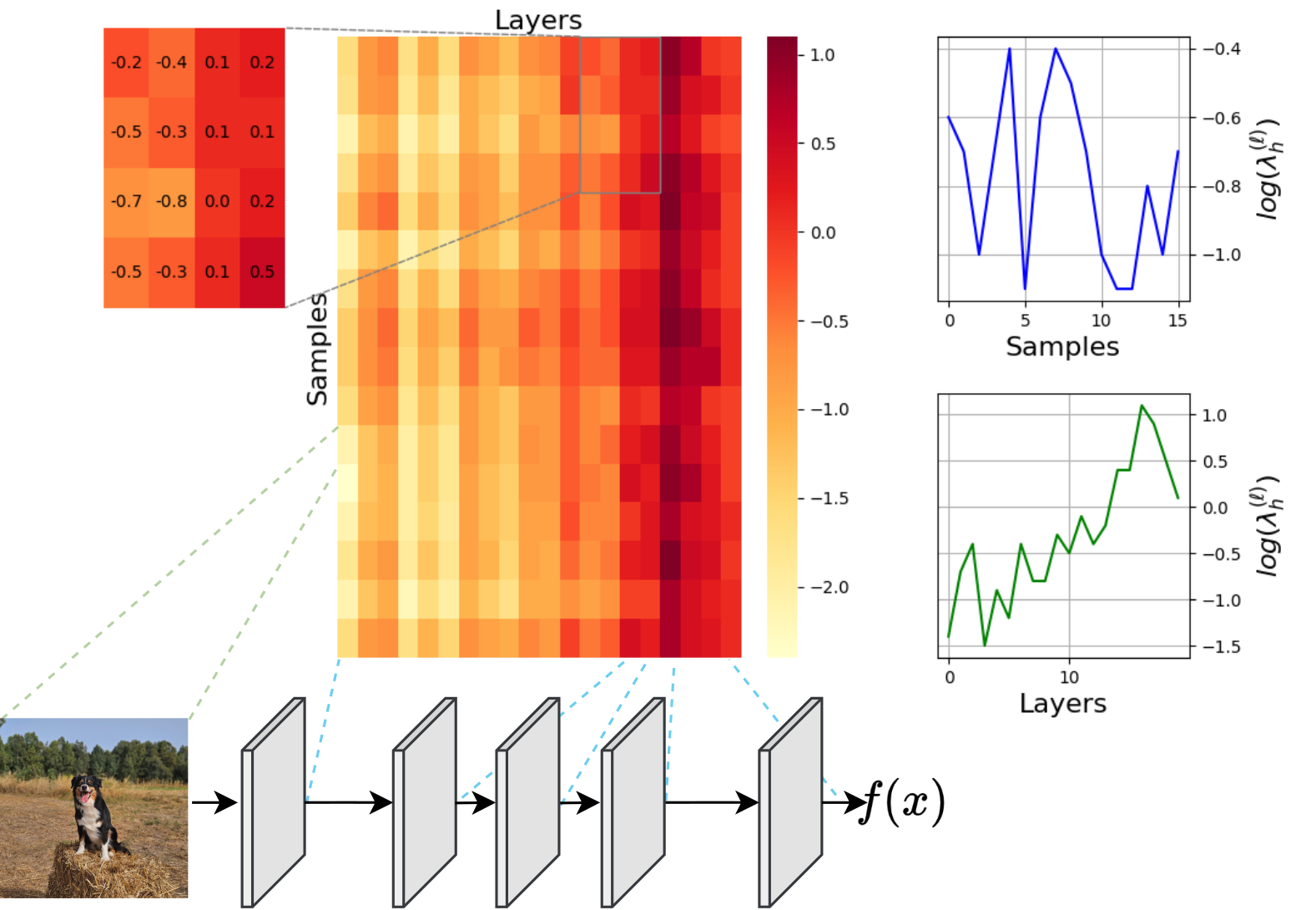}
    \caption{Sample-Layer Attention illustration for network-wise optimization, on ResNet18 with 16 random samples.
    (1) On the \textbf{left}, there is a demonstration of the values of the computed Hessian bound scores w.r.t. layers outputs. 
    The zoomed area shows how various samples yield scores of varying magnitudes across different layers.
    (2) The \textbf{top-right} part presents the scores for different samples for a given layer, showcasing the per-sample attention.
    (3) The \textbf{bottom-right} part presents the scores for all layers for a given sample, showcasing the per-layer attention.
    Note that all values are in log-scale.
    }
    \label{fig:sla-ill}
 \end{figure}

Recently, a new approach for quantization has been suggested in AdaRound \cite{nagel2020up} and was followed by BRECQ~\cite{li2021brecq}, AdaQuant~\cite{hubara2020improving}, QDrop~\cite{wei2021qdrop}, and NWQ~\cite{zheng2022leveraging}.  
These techniques involve a short optimization process that does not rely on labeled data, striking a balance in terms of efficiency and accuracy improvement. 
This optimization process aims to enhance the sub-optimal rounding-to-nearest policy used in basic quantization for quantized weights.
The core idea behind these methods is minimizing the error between the floating-point model and its quantized counterpart. 
In particular, the distinction among these approaches lies in the nature of their optimization processes. 
AdaRound suggested performing layer-by-layer optimization, whereas BRECQ shows that block-by-block optimization can achieve better results, by accounting for cross-layer dependencies within each optimized block.  
A recent attempt in AdaQuant~\cite{hubara2020improving} to optimize the entire network at once, suffers from a high generalization error when applied to a small dataset.
This occurs due to the attempt to optimize a very large number of parameters simultaneously, using a small number of representative samples, which leads to over-fitting, as detailed in BRECQ~\cite{li2021brecq} and NWQ~\cite{zheng2022leveraging}.
NWQ tackled this problem by combining activation regularization and additional data augmentations. 
However, adding augmentation to the dataset limits the generalization of the quantization method to new domains and different tasks.

In this work, we address the generalization problem and suggest \emph{Enhanced Post-Training Quantization ({\name})} for network-wise optimization, to benefit from cross-layer dependencies. 
To overcome the generalization error limitation of network-wise optimization in the case of a small dataset, we suggest a novel objective function that aims to capture the importance of each layer and sample to the optimization process. 
This objective prioritizes the layers where the quantization error significantly affects task performance.

At the core of the objective function, we rely on Hessian matrix analysis, a well-established metric in model compression and optimization studies~\cite{dong2019hawq, dong2020hawq, yao2020pyhessian, yu2022hessian}.  
Notably, while the Hessian matrix effectively captures layer sensitivity, its computation typically requires labeled data, rendering it unsuitable for PTQ.
To overcome this challenge, we suggest a Hessian-based \emph{Sample-Layer Attention (SLA)} score that enables to upper bound the quantization error, allowing the use of label-free Hessian scores. Furthermore, we adapt this methodology to improve the selection of weights quantization parameters. 
Traditional PTQ methods usually rely on the optimization of MSE~\cite{choukroun2019low, nahshan2021loss} or cosine similarity~\cite{wu2020easyquant}, which can yield sub-optimal results~\cite{yuan2022ptq4vit}.
Alternatively, our \emph{Hessian-based MSE} ({\hmse}) method prioritizes minimizing the quantization error for parameters within the weight tensor that have more impact on the task loss. 

In our experiments, we present results on three tasks (Image Classification, Object Detection, and Semantic Segmentation) and on various neural network architectures that show state-of-the-art performance in most cases. 
These results are indicative of the efficacy of {\name}.
For example, we improve the results for MobileNetV2 weights and activation quantization with a low bit-width by more than $7\%$ in certain settings.
In addition, we perform an ablation study to analyze the effect of each component of {\name}, as well as to show the robustness of various dataset sizes and number of optimization iterations.  

In particular, our contributions are the following:
\begin{enumerate}

    \item We propose an optimization method for PTQ, named {\name}. 
    It is model-agnostic and based on knowledge distillation loss with a sample-layer attention score. The attention score guides the optimization towards compensating for the accuracy degradation caused by weights perturbation. 
    
    \item We introduce a Hessian-based MSE ({\hmse}) metric that improves weights quantization parameters selection by minimizing the quantization error on the more sensitive element in the weights tensor.

    \item We present state-of-the-art results using {\name} for a wide variety of tasks (Image Classification, Semantic Segmentation, and Object Detection) and datasets (ImageNet, COCO, and Pascal VOC).

\end{enumerate}
In the spirit of reproducible research, the code of this work is available at: \href{https://github.com/ssi-research/eptq-sla}{https://github.com/ssi-research/eptq-sla}.
The algorithm is also available as part of Sony's open-source Model-Compression-Toolkit library at: \\ \href{https://github.com/sony/model_optimization}{https://github.com/sony/model\_optimization}.

%% file: files/background.tex
\section{Related Work}

 
\subsection{Post-Training Quantization Optimization}

In recent research,  there has been notable progress in strategies for PTQ.
Several methods that received a lot of attention are layer-by-layer and block-by-block optimization. 
A layer-by-layer PTQ approach was first introduced in~\adaround. 
This method explores the second-order term in the Taylor series expansion of the task loss, revealing that the widely used rounding-to-nearest quantization method falls short in terms of optimizing the task loss.
Instead, this approach advocates for learning the rounding policy by optimizing a per-layer local loss.

Building upon the AdaRound framework,~{\adaquant} relaxed its implicit constraint that confines quantized weights within ${\pm 1}$ of their round-to-nearest values in addition to performing the optimization per-network. 
Later,~{\brecq} introduced another significant advancement by leveraging the fundamental building blocks in neural networks and systematically reconstructing them one by one. 
This allows a good balance between cross-layer dependency and generalization but is limited to account only for inner-block layer dependencies.

The next major leap was achieved in~\qdrop, 
which introduced a pivotal adjustment during retraining by incorporating random drops of activation quantization.
This addition enables the model's refined weights to adapt to activation quantization.
Recent works, like {\nwq}, have refined QDrop's random activation drop policy by employing an annealing process to the quantization drop probability. 
This modification, integral to their network-wise optimization technique, has proven beneficial. 
Additionally, they introduced activation regularization loss to address the limitations of network-wise optimization.

In parallel explorations, other works addressed different aspects of the retraining process that may affect the quantized model results.
Ma et al.~\cite{ma2023solving} analyzed the oscillation problem that appears to be caused by the per-block optimization methodology.
{\pdquant} introduced the Prediction Difference metric to dynamically adjust quantization parameters during retraining, contributing to an enhanced outcome in the overall optimization process.


\subsection{Hessian for Model Compression}
Most model compression methods, such as quantization and pruning, 
require some assessment of the sensitivity of the network's layers or blocks to quantization.
One popular method, originally introduced in HAWQ \cite{dong2019hawq}, is to use second-order information to assess the quantization sensitivity of each block.
The authors suggest computing the eigenvalues of the Hessian matrix of the model's task loss w.r.t. the parameters of each block, to use in mixed-precision quantization. 
Q-BERT~\cite{shen2020q} adopts this idea and applies it to a transformer-based architecture to quantize its parameters to a very low precision.
HAWQ-V2~\cite{dong2020hawq} and HAWQ-V3~\cite{yao2021hawq} further extend the idea and use the trace of the Hessian matrix to assess the sensitivity of each block. 

It is worth mentioning that all aforementioned works are computing the Hessian information and using it in a QAT framework, in which the dataset labels and the models' training loss function are available.
Nonetheless, PTQ optimization methods like~{\adaround} and~{\brecq} also made a comprehensive theoretical study of the second-order error of the loss induced by quantization.
They then use their analysis to improve the weights optimization process. Yet, these methods do not compute or use the Hessian information as part of the optimization.
A similar idea is also used for structured pruning in different works~\cite{hassibi1992second, dong2017learning, yu2022hessian}.


%% file: files/preliminaries.tex
\section{Preliminaries}

\paragraph{Notations.}
Throughout this work, vectors are denoted with small bold letters $\vectorsym{a}$, and matrices are denoted with capital bold letters $\matsym{A}$.
The~${i^{th}}$ element of a vector $\vectorsym{v}$ is denoted by $\vectorsym{v}_i$.
To indicate the $i^{th}$ element of a vector that is the result of a function $f\brackets{x}$ we use $\squareb{f\brackets{x}}_i$.
We use $\mathbb{E}_{\x}$ to refer to the expectation w.r.t. a random vector~$\x$.
For a vector $\vectorsym{v}$ we use $\norm{\vectorsym{v}}_2\triangleq \sqrt{\sum_{i}\vectorsym{v}_i^2}$ to denote the $L_2$ norm of $\vectorsym{v}$.
For a matrix~${\matsym{A}}$ we use $\vectorsym{a}=\text{Diag}\brackets{\matsym{A}}$ to represent the vector which is the diagonal of~$\matsym{A}$.

\paragraph{Network and Loss function.}
We denote the floating point neural network by ${f:\mathbb{R}^{d_i}\xrightarrow{}\mathbb{R}^{d_o}}$ and consider $f$ to have $L$ layers. 
The parameters of $f$ are denoted with $\wf = \{\idxw{1}, \idxw{2}, \ldots, \idxw{L}\}$, where $\lw$ are the parameters of layer $\layer$.
Respectively, the activation tensor of layer $\layer$ is denoted with $\lz$.
The network $f$ is optimized for a task loss ${\mathcal{L}_{task}}:(\mathbb{R}^{d_o},\mathbb{R}^{d_o})\xrightarrow{}\mathbb{R}$.
The input and label vectors of the network are denoted by $\x\in\mathbb{R}^{d_i}$ and~${\y\in\mathbb{R}^{d_o}}$, respectively.


\paragraph{Hessian matrix approximation.}
Previous works have shown that the loss degradation of the network due to quantization can be approximated using the Hessian matrix of the task loss w.r.t. the activation tensors of the network \cite{nagel2020up, li2021brecq}.
Thus, computing the Hessian matrix becomes a key component in utilizing second-order derivative information for approximating the effect of quantization on the network's loss.
Unfortunately, computing the Hessian matrix is infeasible due to the very large size of the matrix.
Therefore, approximating the relevant Hessian information is needed.

Following the network $f$ notations, the Hessian of the task loss w.r.t. a tensor~${\zf\in\mathbb{R}^{d_k}}$ is denoted as~$\hessian{\zf}\in\mathbb{R}^{d_k\times d_k}$ and is given by:
\begin{equation}
\label{eq:hessian-matrix}
   {\hessian{\zf}\brackets{\x,\y}}\triangleq {\frac{\partial^2\mathcal{L}_{task}\brackets{\y,f\brackets{\x}}}{\partial\zf\partial\zf^T}}.
\end{equation}
We follow the same process from previous works~\cite{dong2020hawq, nagel2020up, li2021brecq},
using the Hessian matrix definition and the chain rule as detailed in Appendix~\ref{appx:hessian-process}.
Assuming that the pre-trained network's weights, $\wf$, have converged to minimize the task loss, then the gradients are close to~0~\cite{yao2021hawq, nagel2020up, li2021brecq}, and the Hessian approximates the Gauss-Newton matrix~\cite{botev2017practical}.
This results in the following approximation of the Hessian matrix:
\begin{equation}
\label{eq:hessian-approx}
\hessian{\zf}\brackets{\x,\y}\approx\matsym{J}^{(\zf)}\brackets{\x}^T\at{\matsym{B}\brackets{\y,\vectorsym{r}}}{\vectorsym{r}=f\brackets{\x}}\matsym{J}^{(\zf)}\brackets{\x},
\end{equation}
where~${\matsym{B}\brackets{\y,\vectorsym{r}}= \frac{\partial^2\mathcal{L}_{task}\brackets{\y,\vectorsym{r}}}{\partial\vectorsym{r}\partial\vectorsym{r}^T}}$
is the Hessian of the task loss w.r.t. the network's output and $\matsym{J}^{(\zf)}$ is the Jacobian matrix w.r.t. $\zf$ . 
Note that, to compute $\matsym{B}$, the network's task loss function is required, along with a labeled dataset.

%% file: files/method.tex
\section{Method}
\label{sec:method}

In this section, we introduce {\name}, an enhanced PTQ method that optimizes the quantized weights rounding policy on the entire network at once, to account for cross-layer dependencies during the reconstruction.
To overcome the generalization challenge in network-wise optimization, we propose a novel sample-layer attention score that upper bounds the task loss variation due to weights perturbation.
This is based on the insight that the Hessian matrix of the task loss can be upper-bounded without requiring labels and any knowledge of the task loss.
Based on this observation, we also introduce an enhanced metric for selecting weights quantization parameters, named {\hmse}.


Throughout this section, we refer to a flattened quantized weights tensor $\lw$ and activation tensor~$\lz$ of layer~$\layer$ by~$\quantlw$ and~$\quantlz$, respectively.
Our goal is to optimize the weights quantization $\Delta\wf\triangleq\wf-\tilde{\wf}$ to minimize the change in the task loss caused by the perturbation, that is:
\begin{equation}
    \label{eq:weight-pert}
    \lossdiff=
    \expectation{\mathcal{L}_{task}\brackets{\y,f\brackets{\x;\wf+\Delta\wf}}}{\y,\x}-\expectation{\mathcal{L}_{task}\brackets{\y,f\brackets{\x;\wf}}}{\y,\x}.
\end{equation}

We start by introducing the \emph{Label-Free Hessian (LFH)} approximation in Proposition~\ref{prop:lfh}.  
The proposition shows that the Hessian matrix of the task loss can be upper-bounded in a label-free manner, which makes it suitable for PTQ scheme.
First, we present the following assumption regarding the Hessian of the task loss:
\begin{assumption}
    \label{asm:lfh}
    Assume that the Hessian of the task loss~${\matsym{B}\brackets{\y,\vectorsym{r}}= \frac{\partial^2\mathcal{L}_{task}\brackets{\y,\vectorsym{r}}}{\partial\vectorsym{r}\partial\vectorsym{r}^T}}$ w.r.t. the networks's output $\vectorsym{r}$ is independent of the label vector~$\y$ and therefore can be denoted by $\matsym{A}\brackets{\vectorsym{r}}$. 
\end{assumption}
Note that Assumption~\ref{asm:lfh} holds for many common loss functions like MSE and Cross-Entropy, as illustrated in Appendix~\ref{apx:label_free_hessian}.

Assumption~\ref{asm:lfh} allows us to dispose of the necessity of a labeled dataset when computing the Hessian matrix approximation.
\begin{proposition}[Label-Free Hessian]
\label{prop:lfh}
Assume that a network $f$ minimizes task loss $\mathcal{L}_{task}$ such that the following assumption:~${\at{\nabla_{\vectorsym{r}}\mathcal{L}_{task}\brackets{\y,\vectorsym{r} }}{\vectorsym{r}=f\brackets{\x}} \approx 0}$ is satisfied. 
In addition, given that Assumption~\ref{asm:lfh} holds, we get that: 
\begin{equation}
    \label{eq:lfh}    \hessian{\zf}\brackets{\x}\approx\matsym{J}^{(\zf)}\brackets{\x}^T\at{\matsym{A}\brackets{\vectorsym{r}}}{\vectorsym{r}=f\brackets{\x}}\matsym{J}^{(\zf)}\brackets{\x},
\end{equation}
and if there exists a constant $c>0$ such that $\matsym{A}\brackets{r}\preceq c\matsym{I}$, then:
\begin{equation}\label{eq:lfh_bound}   
    \hessian{\zf}\brackets{\x}\leq c\matsym{J}^{(\zf)}\brackets{\x}^T\matsym{J}^{(\zf)}\brackets{\x}.
\end{equation}
\end{proposition}
The proof of Proposition~\ref{prop:lfh} is by applying Assumption~\ref{asm:lfh} and using Eq.~\ref{eq:hessian-approx}. 
Note that, by utilizing Equation~\ref{eq:hessian-approx} we apply its relevant assumptions.
In \eqref{eq:lfh_bound} we also apply the upper bound assumption $\matsym{A}\brackets{r}\preceq c\matsym{I}$, which holds for common loss functions, as detailed in Appendix~\ref{apx:label_free_hessian}. 
In addition, in Appendix~\ref{appx:hessian-approx-figs} we compare the LFH to the actual Hessian and demonstrate that the upper bound approximates the relation between layers up to a scale $c$.

Our method relies on Proposition~\ref{prop:lfh} to construct an objective loss function for implementing the two assets of {\name}: (1) A sample-layer attention score for improved weights quantization rounding optimization; and (2) a weights perturbation bound for an improved weights quantization parameters selection.

\subsection{Sample-Layer Attention Bound}
\label{sec:sample-layer-attn}

A previous work~\cite{nagel2020up} showed the connection between the change in task loss caused by weight perturbation with the Hessian matrix of task loss w.r.t. the activations. 
This connection holds under several standard assumptions~\cite{nagel2020up,li2021brecq} that the network $f$ minimizes task loss $\mathcal{L}_{task}$ and that the Hessian matrix w.r.t. the layer $\layer$'s activation tensor $\hessian{\lz}$ is diagonal for each $\layer \in L$, and is given by:
\begin{equation}
    \label{eq:adaround-res}
     \lossdiff
    \approx\sum_{\layer}\sum_{\chan} \expectation{\hessian{\lz}_{\chan,\chan}\brackets{\idxz{\layer}_\chan-\hat{\zf}^{(\layer)}_\chan }^2}{\y,\x},
\end{equation}
where $\hat{\zf}^{(\layer)}$ is the output of layer $\layer$ when only the weights of that layer are quantized. 
Note that \cite{nagel2020up} additional assumed that $\hessian{\lz}_{\chan,\chan}$ is equal to some constant $c_{\chan}$. 

Here, we relax this assumption by introducing a \emph{Sample-Layer Attention (SLA)} objective that gives a tighter upper bound to the change in the task loss.
We follow~\eqref{eq:adaround-res} and Proposition~\ref{prop:lfh}. In addition, we assume that there exists a constant $c>0$ such that $\matsym{A}\brackets{r}\preceq c\matsym{I}$, and derive the following result:
\begin{align}
    \label{eq:sample-layer-attn}
    \lossdiff \leq c\sum_{\layer}^L \expectation{\maxhess{\layer}\brackets{\x}\cdot\norm{\lz-\hat{\zf}^{(\ell)}}_2^2}{\x},
\end{align}
where $\maxhess{\layer}\brackets{\x}=\max\limits_{i}\squarejac^{\brackets{\lz}}_i$ is the sample-layer attention score and $\squarejac^{\brackets{\lz}} = {\text{\textup{Diag}}\brackets{\matsym{J}^{(\lz)}\brackets{\x}^T\matsym{J}^{(\lz)}\brackets{\x}}}$ is an upper bound on the Hessian diagonal elements. 
The details of the derivation of the result in Equation~\ref{eq:sample-layer-attn} are given in Appendix~\ref{apx:sample-attention-proof}. 
The factor,~$\maxhess{\layer}$, is computed per sample~$\x$ and provides relative attention to each layer~$\layer$'s error on each sample in the representative dataset.
In practice, it is computed using a modification of the Hutchinson algorithm \cite{yao2020pyhessian, avron2011randomized} for a fast approximation of the Jacobian matrix (Appendix~\ref{apx:efficient_jac}). 
This modification eliminates the need for computing the Jacobian matrix, which can be very costly.

The per-sample per-layer attention is illustrated in Figure~\ref{fig:sla-ill}.
The heatmap in this figure presents the magnitude variation between the scores provided by different samples to each layer.
We observe, highlighted in the zoomed region, that different layers get stronger values for different samples and, hence, are more impactful for the network's outcome for that specific sample.
This evidence supports the motivation for computing a sample-layer score.
In addition, we can see that some layers, mainly the last layers in the network, would usually get greater attention for most of the samples, making it important to guide the optimization to focus on them.
The outcome is an objective function that directs the optimization to the most influential samples and layers, thereby enhancing the optimization of the quantized network, as shown empirically in Table~\ref{tab:sla-ablation}.

\subsection{Hessian-based Mean Squared Error}
\label{sec:hmse}
We follow the observation in Proposition~\ref{prop:lfh} to introduce an improved metric to select weight quantization parameters.
To achieve this, we bound the change in task loss caused by the weight perturbation.
We present \emph{Weights Perturbation Bound} in \eqref{eq:hmse-loss} and introduce a new optimization objective called \emph{Hessian-based Mean Squared Error} (\hmse).
This objective is later employed in our proposed PTQ framework to improve quantization parameters selection. 
For this end, we follow the standard assumption \cite{nagel2020up,li2021brecq} that the network $f$ minimizes task loss $\mathcal{L}_{task}$ resulting in $\expectation{{\at{\nabla_{\vectorsym{r}}\mathcal{L}_{task}\brackets{\y,\vectorsym{r} }}{\vectorsym{r}=f\brackets{\x}}}}{\x, \y} \approx 0$, which lead to the following approximation:
\begin{equation}
\lossdiff_{\lw} \approx\brackets{\Delta\lw}^T\matsym{H}^{\brackets{\lw}} \Delta\lw, 
\end{equation}
where $\matsym{H}^{\brackets{\lw}}$ is the Hessian matrix w.r.t. the layer $\layer$'s weights tensor.

To employ a label-free computation of $\matsym{H}^{\brackets{\lw}}$, we assume that there exists a constant $c>0$ such that $\matsym{A}\brackets{r}\preceq c\matsym{I}$. 
Furthermore, for efficiency computation, we assume that $\hessian{\lw}$ is diagonal for each $\layer \in L$. This results in the following upper bound:  
    \begin{equation}
        \label{eq:hmse-loss}
        \lossdiff_{\lw}\leq c\cdot\textnormal{{\hmse}}\brackets{\lw,\quantlw} \triangleq c\sum_{i} \vectorsym{h}^{(\layer)}_{i} \brackets{\lw_i - \quantlw_i}^2,
    \end{equation}
where $\vectorsym{h}^{(\layer)}\triangleq \text{\textup{Diag}}\brackets{\expectation{\matsym{J}^{(\lw)}\brackets{\x}^T\matsym{J}^{(\lw)}\brackets{\x}}{\x}}.$ A detailed derivation of the bound in \eqref{eq:hmse-loss} is shown in Appendix~\ref{apx:proof_hmse}. 
This bound using {\hmse} is tighter than the one achieved using a basic MSE, which allows for an improved search process of quantization parameters, as shown empirically in Table~\ref{tab:hmse}.

Unlike the result from Equation~\ref{eq:sample-layer-attn}, which bounds the change in task loss using a per-layer, per-sample attention score, the parameters selection is computed independently for each layer, making a per-layer score irrelevant. 
Instead, we aim to account for the impact that each element in the weights tensor has on the task loss. The value of~${\vectorsym{h}^{(\layer)}}$ in~\eqref{eq:hmse-loss} provides each element of the weights vector with an attention score, indicating that impact.

\subsection{Enhanced PTQ (EPTQ) Process}
\label{sec:eptq}
We combine all mentioned components to construct an \emph{Enhanced Post-Training Quantization (\name)} method.
It starts by finding improved quantization parameters using {\hmse}~(Section~\ref{sec:hmse}) 
Then, {\name} utilizes the Sample-Layer Attention from~\ref{eq:sample-layer-attn} to perform a network-wise optimization to regain accuracy by improving the rounding of the quantized weight values.
The entire EPTQ workflow is depicted in the algorithm in Appendix~\ref{appx:workflow}.

\paragraph{Threshold Selection.}
\label{sec:threshold}
Using the bound in \eqref{eq:hmse-loss}, we propose a new weights quantization parameters selection minimization.
In this work, the weights are quantized using a symmetric quantizer, that is, we focus on finding an optimized threshold $t^{(\layer)}$ for quantizing the weights of each layer $\layer$.
First, we obtain the Hessian diagonal $\vectorsym{h}^{(\layer)}$ using LFH, as explained in Section~\ref{sec:hmse}. 
Then, we perform a threshold search for each layer weights $\lw$ using the following minimization objective: 
\begin{equation}
    t^{(\layer)} = \arg\min\limits_{t}\textnormal{\hmse}\brackets{\lw,\quantlw\brackets{t}},  
\end{equation}
where $\quantlw\brackets{t}$ denotes the quantized tensor $\quantlw$ with threshold $t$.

\paragraph{Hessian-aware Network-wise optimization loss.}
\label{sec:optimization}
We use the symmetric quantizer from~\cite{nagel2020up}, which updates the rounding value $\vectorsym{v}$ of the quantized weights between the floor and ceiling points.
Using SLA~\eqref{eq:sample-layer-attn} we construct a Hessian-aware knowledge distillation loss to transfer knowledge from the floating-point network to the quantized network~\cite{hinton2015kd, mishra2017apprentice, polino2018model, yao2021hawq}. 
We utilize the sample-layer attention for more sensitive layers, to improve the optimization process and overcome the network-wise reconstruction limitations.
This allows us to benefit from considering cross-layer dependencies during optimization.
To enable this, we replace $\hat{\zf}^{(\layer)}$ by $\quantlz\brackets{\vectorsym{v}}$, which includes the quantization error of all previous layers. 
This results in a knowledge-distillation optimization loss that is defined as follows:
\begin{equation}
\begin{alignedat}{1}
\label{eq:kd-loss}
    \kdmInput{\vectorsym{v}} = \sum_{\layer}^L \expectation{\maxhess{\layer}\brackets{\x}\cdot\norm{\lz - \quantlz\brackets{\vectorsym{v}}}_2^2}{\x} + \lambda_{reg}\cdot f_{reg}(\vectorsym{v}),    
\end{alignedat}
\end{equation}
where $f_{reg}\brackets{\vectorsym{v}}$ is the regularization factor that encourages the optimization to converge towards rounding the quantized values either up or down and $\lambda_{reg}$ is a hyper-parameter that determines the rate of converge, similar to~\cite{nagel2020up}. 


\paragraph{Gradual Activation Quantization.}

We follow the annealing activation quantization drop approach from~\cite{zheng2022leveraging} and apply \emph{Gradual Activation Quantization} during the optimization process.
The difference in our policy is that it does not include any randomization, making the optimization process more stable.
Specifically, let $\mathcal{P}^{(\layer)}\brackets{i}$ be the percentage of quantized activations in the output tensor of layer $\layer$ in the $i^{th}$ optimization iteration.
Then, the quantized activation output of a layer $\layer$ in the $i^{th}$ iteration is defined as follows:
\begin{equation}
    \quantlz = \mathcal{P}^{(\layer)}\brackets{i} \cdot \x^{(\layer)} + \brackets{1 - \mathcal{P}^{(\layer)}\brackets{i}} \cdot Q\brackets{\x^{(\layer)}},
\end{equation}
where $\x^{(\layer)}$ is the floating-point activation output of layer $\layer$ and $Q\brackets{\x^{(\layer)}}$ is the quantized tensor $\x^{(\layer)}$.
We set $\mathcal{P}^{(\layer)}\brackets{0}$ to some pre-defined initial percentage, and reduce it using linear decay during optimization until it reaches zero, i.e., the entire tensor is quantized.

\subsection{Discussion}

While {\name} can be applied to various architectures and tasks, several limitations should be noted. 
First, SLA~(\ref{sec:sample-layer-attn}) and {\hmse}~(\ref{sec:hmse}) 
rely on bounds that utilize the LFH proposition~(\ref{prop:lfh}) which assumes Assumption~\ref{asm:lfh}.
These bounds may not accurately approximate loss functions that do not meet this assumption, such as Sharpness-Aware Maximization (SAM)~\cite{foretsharpness}. 
Second, the computational cost is derived from Hessian approximation computation and the number of optimization iterations. 
For instance, running the algorithm with 80K steps on a ResNet18 using an NVIDIA GeForce RTX 3090 GPU takes about 4 hours, with runtime increasing for larger models. 
Lastly, the global optimization approach may be more sensitive to hyper-parameter tuning than other local methods.

%% file: files/experimental.tex
\section{Experimental Results}
\label{sec:experimets}

In this section, we conduct extensive numerical experiments to validate {\name} and its different components (Section~\ref{sec:eptq}).
We demonstrate that employing the Hessian-aware objective loss frequently achieves state-of-the-art results for various CNN architectures and diverse tasks.

\subsection{Experimental Settings}
Our comparison is based on the published results of other optimization-based PTQ methods, such as \brecq, \qdrop, and \pdquant.
First, we apply basic quantization and incorporate the {\hmse} for the per-channel threshold selection in the symmetric weights quantization, as detailed in Section~\ref{sec:eptq}.
Then, we apply the Hessian-guided EPTQ optimization to recover the accuracy lost due to the quantized weights rounding policy.
Additionally, we optimize the quantization scaling factors and the biases of linear operations.
Additional specifications of our PTQ scheme are depicted in Appendix~\ref{appx:quant_setup}.

We run EPTQ with a representative dataset of 1,024 i.i.d. samples for 80K gradient steps. 
We report the mean and standard deviation of five runs with different initial seeds for each experiment.
The full details regarding the optimization setting and hyper-parameters can be found in Appendix~\ref{apx:opt-setup}.
The experiments are conducted on models provided by {\brecq}. 
Finally, we set the first and last layer quantization to 8-bit for all experiments, similar to all compared works.
All experiments are conducted on a single NVIDIA GeForce RTX 3090 GPU.

\subsection{Accuracy Comparison}

We compare the results of {\name} on different networks for the ImageNet classification task, against the reported results of multiple other PTQ algorithms across different bit-width settings. 
We ensure that our settings align with those of the compared works, resulting in a comprehensive set of experiments.
The results are presented in Table~\ref{tab:all-res}.
Additional results, comparing EPTQ with the work of Ma et al.~\cite{ma2023solving}, which uses a larger representative dataset, are presented in Appendix~\ref{apx:add-res}.

In summary, it can be observed that, because of the incorporation of network-wise optimization and SLA, {\name} outperforms the other layer-wise and block-wise by a noticeable margin in the majority of tested settings and on various networks.
For example, with MobileNetV2 using 3-bit weight and activation quantization, {\name} achieves an accuracy improvement of over $7\%$ compared to the block-wise approach of QDrop.
When compared to NWQ, which employs data augmentations in its network-wise optimization approach to address generalization issues, {\name} demonstrates superior results in most cases and competitive results in others, all without the need for any data augmentations.

\begin{table}[tb]
\caption{Comparison between {\name} and different PTQ methods on ImageNet classification task. * means using BRECQ's setting of keeping the first layer output in 8-bit. $\dagger$ represents methods that utilize additional data augmentations in their experiments.}
\label{tab:all-res}
\centering
\resizebox{0.98\textwidth}{!}{%
\begin{tabular}{lccccccc}
\hline
\textbf{Methods} & \textbf{Bits (W/A)} & \textbf{ResNet18} & \textbf{ResNet50} & \textbf{MobileNetV2} & \textbf{RegNet-600M} & \textbf{RegNet-3.2GF} & \textbf{MnasNet-2.0} \\ \hline
Full Prec.       & 32/32               & 71.01             & 76.63             & 72.62                & 73.52                & 78.46                 & 76.52                \\ \hline
AdaRound         & \multirow{3}{*}{4/32}                & 68.71             & 75.23             & 69.78                & 71.97                & 77.12                 & 74.87                \\
BRECQ            &                 & 70.70             & 76.29             & 71.66                & 73.02                & 78.04                 & 76.00                \\
\textbf{\our}     &                 & \textbf{70.75} \tstd{0.09}             & \textbf{76.46} \tstd{0.07}             & \textbf{72.06} \tstd{0.04}               & \textbf{73.11} \tstd{0.09}               & \textbf{78.16} \tstd{0.06}                    & \textbf{76.02} \tstd{0.13}                    \\ \hline
AdaRound         & \multirow{3}{*}{3/32}                & 68.07             & 73.42             & 64.33                & 67.71                & 72.31                 & 69.33                \\
BRECQ            &                 & 69.81             & 75.61             & 69.50                & 71.48                & 77.22                 & 74.58                \\
\textbf{\our}     &                 & \textbf{70.19} \tstd{0.03}             & \textbf{75.72} \tstd{0.06}             & \textbf{70.04} \tstd{0.35}                & \textbf{71.85} \tstd{0.11}                & \textbf{77.41} \tstd{0.07}                     & \textbf{75.20} \tstd{0.27}                   \\ \hline
AdaRound         & \multirow{3}{*}{4/8}                 & 68.55             & 75.01             & 69.25                & -                    & -                     & -                    \\
BRECQ            &                  & 70.58             & 76.29             & 71.42                & 72.73                & -                     & -                    \\
\textbf{\our}     &                 & \textbf{70.70}             & \textbf{76.45}             & \textbf{72.06} \tstd{0.04}               & \textbf{73.10} \tstd{0.03}               & \textbf{78.13} \tstd{0.05}                     & \textbf{76.01} \tstd{0.13}                   \\ \hline
AdaRound         & \multirow{4}{*}{4/4}                 & 67.96             & 73.88             & 61.52                & 68.20                & 73.85                 & 68.86                \\
QDROP            &                & 69.17             & 75.15             & 68.07                & 70.91                & 76.40                 & 72.81                \\
PD-Quant         &                 & 69.23             & 75.16             & 68.19                & 70.95                & 76.65                 & 73.26                \\
\textbf{\our}     &                 & \textbf{69.52} \tstd{0.09}            & \textbf{75.45} \tstd{0.08}             & \textbf{69.96} \tstd{0.06}               & \textbf{71.58} \tstd{0.14}                & \textbf{76.95} \tstd{0.09}                 & \textbf{74.52} \tstd{0.07}                \\ \hdashline
AdaRound$^{*}$         & \multirow{6}{*}{4/4}                 & 69.36             & 74.76             & 64.33                & -                    & -                     & -                    \\
BRECQ$^{*}$            &                 & 69.60             & 75.05             & 66.57                & 68.33                & 74.21                 & 73.56                \\
QDROP$^{*}$            &                 & 69.62             & 75.45             & 68.84                & 71.18                & 76.66                 & 73.71                \\
PD-Quant$^{*}$         &                 & 69.72             & -                 & 68.76                & -                    & -                     & -                    \\
NWQ$^{*\dagger}$              &                 & 69.85             & -                 & 69.14                & \textbf{71.92}                & \textbf{77.40}                 & 74.60                \\
\textbf{\our}$^{*}$     &                 & \textbf{69.91} \tstd{0.08}            & \textbf{75.71} \tstd{0.06}            & \textbf{70.14} \tstd{0.06}                & 71.86 \tstd{0.04}                & 77.15 \tstd{0.09}                 & \textbf{74.76} \tstd{0.13}               \\ \hline
QDROP            & \multirow{2}{*}{3/3}                 & 65.65             & 71.29             & 54.59                & 64.57                & 71.73                 & 63.83                \\
\textbf{\our}     &                  & \textbf{66.87} \tstd{0.11}            & \textbf{72.37} \tstd{0.10}            & \textbf{62.03} \tstd{0.05}                & \textbf{66.59} \tstd{0.16}               & \textbf{73.55} \tstd{0.13}                & \textbf{70.38} \tstd{0.17}               \\ \hdashline
AdaRound$^{*}$         & \multirow{5}{*}{3/3}                 & 64.66             & 66.66             & 15.20                & 51.01                & 56.79                 & 47.89                \\
BRECQ$^{*}$            &                 & 65.87             & 68.96             & 23.41                & 55.16                & 57.12                 & 49.78                \\
QDROP$^{*}$            &                 & 66.75             & 72.38             & 57.98                & 65.54                & 72.51                 & 66.81                \\
NWQ$^{*\dagger}$              &                 & 67.58             & -                 & 61.24                & 67.38                & \textbf{74.79}                 & 68.85                \\
\textbf{\our}$^{*}$     &                 & \textbf{67.84} \tstd{0.09}             & \textbf{73.29} \tstd{0.10}             & \textbf{63.05} \tstd{0.27}               & \textbf{67.58} \tstd{0.08}               & 74.23 \tstd{0.07}                & \textbf{71.10} \tstd{0.27}                \\ \hline
\end{tabular}
}
\end{table}

\subsubsection{Additional Tasks}
To demonstrate the benefits and robustness of {\name}, we experimented with it on two additional tasks--semantic segmentation and object detection, showing our approach excels beyond image classification.

\paragraph{Semantic Segmentation.}
Table~\ref{tab:segmentation} presents a comparison between {\name} and other quantization
methods on the DeeplabV3+~\cite{chen2018deeplab} model with MobileNetV2 backbone for the semantic segmentation task. 
Note that since the compared floating-point models differ in their baseline accuracy, we present and highlight the relative accuracy degradation ($\resdelta$) of the quantized models, compared with each float baseline.
In this experiment, we run only 20K gradient-step iterations and set the regularization factor to be similar to the MobileNetV2 experiment, as detailed in Appendix~\ref{apx:opt-setup}.
We can observe that the relative accuracy is improved compared to the other results for this task as well.

\paragraph{Object Detection}
Table~\ref{tab:retinanet} shows competitive results with BRECQ quantizing the RetinaNet~\cite{lin2017focal} network with ResNet50 backbone in different bit-width precision while removing the requirement of manually selecting the layers on which we perform the optimization.
Note that in this experiment we run 20K gradient-step iterations.
We also follow BRECQ's methodology of quantizing only the network's backbone and keeping its first and last layers at 8-bit.
It can be seen that {\name} achieves competitive results, without needing to manipulate the model to meet BRECQ's block structure.

\begin{table}[t]
\centering
\begin{minipage}[t]{0.47\textwidth}
    \centering
    \caption{Mean Intersection Over Union results for semantic segmentation on DeeplabV3+~\cite{chen2018deeplab} for 4-bit weight and 8-bit activation.}
    \label{tab:segmentation}
    \resizebox{\textwidth}{!}{%
    \begin{tabular}{@{}lccc@{}}
    \toprule
    \multicolumn{1}{c}{\textbf{Method}}                      & \textbf{Full Prec.}    & \textbf{Q}     & \textbf{$\resdelta$}      \\ \midrule
    DFQ \cite{nagel2019dfq}                        & 72.94 & 14.45 & 58.49         \\
    HPTQ \cite{habi2021hptq}                                        & 75.56 &    56.71   &  18.85             \\
    \adaround                                                & 72.94 & 70.86 & 2.08          \\
    \textbf{\our}                                               & 77.72 & \textbf{76.80} \tstd{0.12} & \textbf{0.94} \\ \bottomrule
    \end{tabular}
    }
\end{minipage}%
\hfill
\begin{minipage}[t]{0.47\textwidth}
    \centering
    \caption{Mean Average Precision results of {\name} and {\brecq} for object detection on RetinaNet~\cite{lin2017focal}. 
    }
    \label{tab:retinanet}
    \resizebox{\textwidth}{!}{%
    \begin{tabular}{lclc}
    \toprule
    \multicolumn{1}{c}{\textbf{Method}} & \textbf{Bits (W/A)}         & \multicolumn{1}{c}{\textbf{Full Prec.}} & \textbf{Q}                         \\ \midrule
    BRECQ                               & \multirow{2}{*}{8/8} & \multirow{2}{*}{36.8}           & 36.73                              \\
    \textbf{\our}        &                      &                                 & \textbf{36.74} \tstd{0.01}                    \\ \midrule
    BRECQ                               & \multirow{2}{*}{4/8} & \multirow{2}{*}{36.8}           & \textbf{36.65}                     \\
    \textbf{\our}        &                      &                                 & 36.54  \tstd{0.01}                            \\ \midrule
    BRECQ                               & \multirow{2}{*}{4/4} & \multirow{2}{*}{36.8}           & \multicolumn{1}{l}{33.47}          \\
    \textbf{\our}        &                      &                                 & \multicolumn{1}{l}{\textbf{34.10}} \tstd{0.01} \\ \bottomrule
    \end{tabular}
    }
\end{minipage}
\end{table}

\subsection{Ablation Study}

We conduct an ablation study of the various components of our proposed method, demonstrating how each of them affects the accuracy of the quantized model. 
Additionally, we investigate the impact of different numbers of samples and optimization iterations (gradient steps) on the process. 
Our findings show that our choice of using 1,024 samples and 80K iterations is not only sufficient but also pushes the potential outcome to the limit.
The experimental setup of all ablations aligns with the specifications outlined in Section~\ref{sec:experimets} unless stated otherwise.

\begin{table}[t]
\centering
\caption{
{\hmse} ablation, showing the improvement to the Top-1 accuracy achieved by {\hmse} on several classification networks in a simple PTQ-only scenario.}
\label{tab:hmse}
\resizebox{0.88\textwidth}{!}{
\begin{tabular}{ccccccc}
\hline
\textbf{Bits (W/A)}                & \multicolumn{3}{c}{\textbf{4/32}}                                  & \multicolumn{3}{c}{\textbf{3/32}}                                  \\ \hline
\textbf{Model}                     & \textbf{ResNet18} & \textbf{ResNet50} & \textbf{MobileNetV2}       & \textbf{ResNet18} & \textbf{ResNet50} & \textbf{MobileNetV2}       \\ \hline
\multicolumn{1}{c|}{MSE}  & 62.53             & 70.57             & \multicolumn{1}{c|}{67.34} & 15.85             & 21.33             & \multicolumn{1}{c}{19.63} \\
\multicolumn{1}{c|}{\textbf{{\hmse}}} & 63.14             & 73.11             & \multicolumn{1}{c|}{67.97} & 27.80             & 44.45             & \multicolumn{1}{c}{30.62} \\ \hline
\end{tabular}
}
\end{table}

\begin{table}[t]
\centering
\begin{minipage}[t]{0.48\textwidth}
    \centering
    \caption{SLA ablation, showing the improvement of using the SLA objective during optimization, compared to a common average weighting approach.
    }
    \label{tab:sla-ablation}
    \resizebox{\textwidth}{!}{%
    \begin{tabular}{lccc}
    \hline
    \multicolumn{1}{c}{\textbf{Model}}    & \textbf{ResNet18} & \textbf{MobileNetV2} & \textbf{\begin{tabular}[c]{@{}c@{}}RegNet\\ -600M\end{tabular}} \\ \hline
    \multicolumn{1}{l|}{Average} & 70.04             & 64.25                & 71.01                                                           \\
    \multicolumn{1}{l|}{\textbf{SLA}}     & \textbf{70.09}    & \textbf{70.04}       & \textbf{71.58}                                                  \\ \hline
    \end{tabular}
    }
\end{minipage}%
\hfill
\begin{minipage}[t]{0.49\textwidth}
    \centering
   \caption{Activation quantization drop ablation showing the effect of applying our gradual quantization method.}
    \label{tab:qdrop-ablation}
    \resizebox{\textwidth}{!}{%
    \begin{tabular}{lccc}
    \hline
    \multicolumn{1}{c}{\textbf{Model}}                                                            & \textbf{ResNet18} & \textbf{MobileNetV2} & \textbf{\begin{tabular}[c]{@{}c@{}}MNasNet\\ -2.0\end{tabular}} \\ \hline
    \multicolumn{1}{l|}{No drop}                                                         & 66.28             & 61.12                & 69.22                                                           \\ \hdashline
    \multicolumn{1}{l|}{\begin{tabular}[c]{@{}l@{}}Stochastic \\ drop\end{tabular}}      & 66.56             & 62.02                & 70.05                                                           \\ \hdashline
    \multicolumn{1}{l|}{\textbf{\begin{tabular}[c]{@{}l@{}}Gradual \\ quantization\end{tabular}}} & \textbf{66.76}    & \textbf{62.2}        & \textbf{70.25}                                                  \\ \hline
    \end{tabular}
    }
\end{minipage}
\end{table}

\begin{figure}[t]
    \begin{minipage}[t]{0.48\textwidth}
        \centering
        \includegraphics[width=\textwidth]{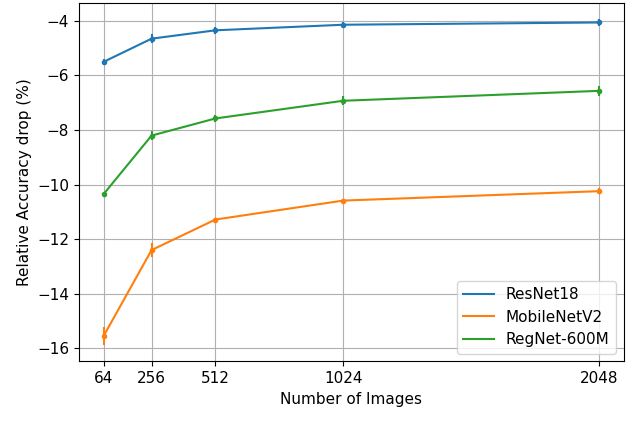}
        \caption{Relative accuracy drop of several classification networks quantized with 3 bits for weights and activation with a different number of images.}
    \label{fig:images-ablation}
    \end{minipage}
    \hfill
    \begin{minipage}[t]{0.48\textwidth}
        \centering
        \includegraphics[width=\textwidth]{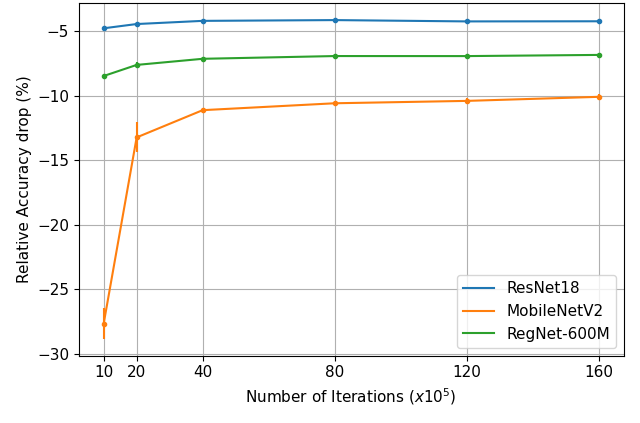}
        \caption{Relative accuracy drop of several classification networks quantized with 3 bits for weights and activation with a different number of iterations.}
        \label{fig:iters-ablation}
    \end{minipage}
\end{figure}

\paragraph{Sample-Layer Attention Ablation.}

Table~\ref{tab:sla-ablation} demonstrates the impact of using the Hessian-guided loss (Section~\ref{sec:optimization}) based on the SLA~\eqref{eq:sample-layer-attn} for the weights quantization rounding optimization.
We illustrate the improvement compared to the common approach of taking simple average weighting, i.e., the factor $\maxhess{\layer}$ is replaced by $\frac{1}{L}$ for all layers.
We measure the accuracy of several networks using the two approaches with weights quantized to 3 bits.
It can be seen that using the SLA-based objective improves the accuracy of all the 
presented networks.

\paragraph{Activation Quantization Drop Ablation.}
We examine the effect of our Gradual Activation Quantization policy in Table~\ref{tab:qdrop-ablation}, compared to the stochastic annealing approach from~\cite{zheng2022leveraging} and a setting without activation quantization drop.
we can see that the benefits from QDrop's activation quantization drop carry over to our new network-wise optimization.
Second, it is notable that our gradual quantization policy further improves the result for all the presented networks, in addition to being more stable, since it removes the randomization element from the process.

\paragraph{\textnormal{{\hmse}} Ablation.}

We showcase the advantages of the objective loss based on our Hessian bound, for selecting weights quantization parameters, termed {\hmse}, over the commonly used MSE method.
In this experiment we ran a basic PTQ method without any additional optimization, to evaluate the sole impact of {\hmse}.
The results, as presented in Table~\ref{tab:hmse}, indicate a significant enhancement with {\hmse}, particularly evident at lower bit-widths compared to the simple MSE method. 
This improvement persists on higher bit-width, enabling decent results with straightforward PTQ without the necessity for additional optimization.

\paragraph{Number of Images.}

Figure~\ref{fig:images-ablation} shows the changes in accuracy due to a change in the size of the dataset.
As can be observed, 
because of the SLA score that factories each layer in the network-wise optimization, 
the relative improvement from increasing the number of images in the representative dataset above 1,024 samples is insignificant.

\paragraph{Number of Iterations.}

Figure~\ref{fig:iters-ablation} shows the effect of the number of optimization iterations on the resulting accuracy.
As can be observed, there is no significant benefit from increasing the number of iterations above 80K, and for some models, it is even sufficient to run only 20-40K iterations.

%% file: files/conclusions.tex
\section{Conclusions}

In this work, we introduced {\name}--a post-training quantization method that consists of a network-wise optimization process for weights quantization rounding reconstruction.
{\name} utilizes knowledge distillation with sample-layer attention score, to overcome the generalization problem of network-wise optimization with a small dataset, allowing it to benefit from accounting for cross-layer dependencies.

We showed that, under conditions that hold for many common loss functions, the Hessian matrix of the task loss can be upper-bounded without the need for labeled data.
This allowed us to leverage the commonly-used approach of using Hessian-based information to assess layers sensitivity to quantization, and construct a novel optimization objective that is suitable for a PTQ scheme.
In addition, we leveraged the LFH observation to propose a new Hessian-based MSE metric for improved weights quantization parameters selection.

We examined {\name} on a large variety of DNN models with different architectures and specifications.
We conducted experiments on three types of problems: ImageNet classification, COCO object detection, and Pascal VOC semantic segmentation. 
We showed that {\name} achieves state-of-the-art results on a wide variety of tasks and models, including ResNet18, ResNet50, and MobileNetV2 for image classification, RetinaNet for object detection, and DeepLabV3+ for semantic segmentation.

%% file: files/ref.tex
\newpage
\bibliographystyle{splncs04}
\bibliography{ref}

%% file: files/appendix.tex
\newpage
\appendix
\input{files/appendix/ptq_details}
\input{files/appendix/proof.tex}
\input{files/appendix/hessian_apendix}

\input{files/appendix/additional_results}

%% file: files/appendix/ptq_details.tex
\section{Post-Training Quantization Scheme}
\label{appx:ptq_deatails}

The overall optimization flow consists of two main stages: (1) Initializing the quantization parameters; (2) Optimizing the rounding parameters.

In the initialization stage, we start with a pre-processing step which consists of folding batch normalization layers into their
preceding convolution layers.
Next, we initialize the quantization threshold $t$ for all channels for each weights tensor $\lw$ for $1 \leq \layer \leq L$
using {\hmse} minimization as detailed in Section~\ref{sec:hmse}.
For this, we first compute the approximated Hessian score w.r.t. the layer weights and then apply the {\hmse} minimization search.

Afterward, we begin the optimization of the network's weights quantization rounding where we first compute the attention score, $\maxhess{\layer}\brackets{\x}$ for each layer $\layer$ and sample $\x$ in the representative dataset.
Then, we use gradient descent to minimize the Hessian-guided knowledge-distillation loss (Equation~\ref{eq:kd-loss}) over the set of parameters $\vectorsym{v}$.
The complete workflow is presented in Algorithm~\ref{alg:workflow}.

\subsection{Quantization Setting}
\label{appx:quant_setup}
In this work,  weights are quantized per-channel using a symmetric quantizer.
We quantize every activation in the neural network with a uniform, per-tensor, quantizer. 
We accept fusion between non-linear and convolution. 
This includes the following quantization points: non-linear, convolution without non-linear, add residual, and concatenation. 
Note that this differs from other methods like BRECQ \cite{li2021brecq}, which only quantizes part of the neural network activations as stated in \cite{jeon2022genie}. 

\subsection{Hyper-parameter and Training Settings}
\label{apx:opt-setup}

For the parameters optimization, we use the RAdam~\cite{liu2019radam} optimizer with default parameters.
Unless stated otherwise, We set the learning rate to 0.01 and the soft quantizer regularization term ($\lambda_{reg}$) to 10, except for MobileNetV2 and MnasNet-2.0 for which $\lambda_{reg} = 20,000$.
We follow the activation quantization parameters from \qdrop, and apply our gradual activation quantization with dropping percentage annealing, followed from~\nwq. We start the percentage annealing at 1.0 for all networks except MobileNetV2 and MnasNet-2.0 which start at 0.5.
For {\hmse}, we use 64 samples to compute the Hessian matrix approximation.

\newpage
\subsection{{\name} Algorithm}
\label{appx:workflow}

Algorithm~\ref{alg:workflow} depicts the workflow of the EPTQ optimization scheme.

\begin{algorithm}[h]
   \caption{{\name} Optimization Workflow}
   \label{alg:workflow}
    \begin{algorithmic}[1]
       \STATE {\bfseries Input:}
       representative dataset $\mathcal{D}_R$, a pre-trained model $f$.
       \STATE Apply Batch Normalization Folding on $f$.
       \STATE Compute LFH approximation w.r.t. weight tensors.
       \STATE Initialize quantization threshold $t$ using {\hmse}~\eqref{eq:hmse-loss}) optimization for each weights quantizer and regular MSE for each activation quantizer.
       \STATE Compute the LFH approximation w.r.t. activation tensors per-samples ${\x\in\mathcal{D}_R}$~\eqref{eq:lfh}.
       \STATE Optimize the quantized weights rounding using SGD minimizing the Hessian-guided knowledge-distillation loss~\eqref{eq:kd-loss} over the dataset $\mathcal{D}_R$.
    \end{algorithmic}
\end{algorithm}

%% file: files/appendix/proof.tex
\section{Proofs}

\subsection{EPTQ Loss Bound}
\label{apx:sample-attention-proof}
For the simplicity of the proofs, we assume that all weighted operations are linear such that $\matsym{W}^{(\ell)}$ and $\tilde{\matsym{W}}^{(\ell)}$ are the floating-point and quantized matrices, respectively. 
We begin with results from AdaRound \cite{nagel2020up} using our notation from Section~\ref{sec:method} and are shown in \eqref{eq:adaround-res}. Now, we assume that the conditions presented in Section~\ref{sec:sample-layer-attn} hold, we present the derivation of the SLA:
\begin{align}\label{eq:proof_eptq_loss}
      \mathcal{L}&\brackets{\wf+\Delta\wf}-\mathcal{L}\brackets{\wf}
    \approx
    \nonumber\\
    &\sum_{\layer}\sum_{\chan}\expectation{\squareb{\matsym{J}^{(\zf)}\brackets{\x}^T\at{\matsym{A}\brackets{\vectorsym{r}}}{\vectorsym{r}=f\brackets{\x}}\matsym{J}^{(\zf)}\brackets{\x}}_{\chan,\chan}\brackets{\zf^{(\layer)}_\chan-\hat{\zf}^{(\layer)}_\chan}^2}{\x}\nonumber\\
    &\leq c\sum_{\layer}\sum_{\chan} \expectation{\squareb{\matsym{J}^{(\zf)}\brackets{\x}^T\matsym{J}^{(\zf)}\brackets{\x}}_{\chan,\chan}\brackets{\zf^{(\layer)}_\chan-\hat{\zf}^{(\layer)}_\chan}^2}{\x}\\\nonumber
    &\leq c\sum_{\layer} \expectation{\maxhess{\layer}\brackets{x}\norm{\zf^{(\layer)}-\hat{\zf}^{(\layer)}}_2^2}{\x},
\end{align}
where $\maxhess{\layer}\brackets{x}=\max\limits_{\chan}\squareb{\matsym{J}^{(\zf)}\brackets{\x}^T\matsym{J}^{(\zf)}\brackets{\x}}_{\chan,\chan}$ is the maximum value of \\$\matsym{J}^{(\zf)}\brackets{\x}^T\matsym{J}^{(\zf)}\brackets{\x}$. In the first step of \eqref{eq:proof_eptq_loss}, we use assumption~\ref{asm:lfh} and \eqref{eq:lfh}. 
Then, we use the assumption that $\matsym{A}\brackets{r}\prec c\matsym{I}$. Finally, we obtain a score per layer by upper bounding $\matsym{J}^{(\zf)}\brackets{\x}^T\matsym{J}^{(\zf)}\brackets{\x}$ using the maximum diagonal value.

\subsection{HMSE Proof}\label{apx:proof_hmse}
Using Taylor series expansion, we have analyzed the perturbation of a single layer weight $\ell$ due to quantization. We begin with:
    \begin{align}
        \mathcal{L}\brackets{\wf^{(\ell)}+\Delta\wf^{(\ell)}}&-\mathcal{L}\brackets{\wf^{(\ell)}}=
        \\ \nonumber
        &\brackets{\Delta\wf^{(\ell)}}^T\vectorsym{g}^{\brackets{\wf^{(\ell)}}}+\brackets{\Delta\wf^{(\ell)}}^T\matsym{H}^{\brackets{\wf^{(\ell)}}} \Delta\wf^{(\ell)} +\mathcal{O}\brackets{\Delta\wf^{(\ell)}},
    \end{align}
    where $\vectorsym{g}^{\brackets{\wf^{(\ell)}}}=\expectation{\nabla_{\wf^{(\ell)}}\mathcal{L}_{task}\brackets{\y,f\brackets{\x;\wf}}}{\x,\y}$ and \\ $\matsym{H}^{\brackets{\wf^{(\ell)}}}=\expectation{\frac{\partial^2\mathcal{L}_{task}\brackets{\y,f\brackets{\x;\wf}}}{\partial\wf^{(\ell)}\partial\brackets{\wf^{(\ell)}}^T}}{\x,\y}$ are gradient and Hessian of task w.r.t. to the weights vector $\wf^{(\ell)}$ of layer $\ell$.
    Using the assumption that the neural network has been well-trained, meaning $\vectorsym{g}^{\brackets{\wf^{(\ell)}}}=0$ we get the following: 
    \begin{align}
        \mathcal{L}\brackets{\wf^{(\ell)}+\Delta\wf^{(\ell)}}-\mathcal{L}\brackets{\wf^{(\ell)}}
        \approx\brackets{\Delta\lw}^T\matsym{H}^{\brackets{\lw}} \Delta\lw.
    \end{align}
    Utilizing Assumption~\ref{asm:lfh} and $\matsym{A}\brackets{\vectorsym{r}} \prec c\matsym{I}$ we have that:

    \begin{align}        \matsym{H}^{\brackets{\lw}}
    &=\expectation{\frac{\partial^2\mathcal{L}_{task}\brackets{\y,f\brackets{\x;\w}}}{\partial\brackets{\lw}\partial\brackets{\lw}^T}}{\x,\y} \\
    \nonumber&=\expectation{\matsym{J}^{\brackets{\lw}}\brackets{\x}^T\at{\matsym{A}\brackets{\vectorsym{r}}}{\vectorsym{r}=f\brackets{\x}}\matsym{J}^{\brackets{\lw}}\brackets{\x}}{\x} \\
    \nonumber&\leq  c 
        \expectation{\matsym{J}^{\brackets{\lw}}\brackets{\x}^T\matsym{J}^{\brackets{\lw}}\brackets{\x}}{\x}.
     \end{align}
Finally, using the assumption that $\matsym{H}^{\brackets{\lw}}$ is diagonal we have that:
\begin{equation}
    \lossdiff\brackets{\wf^{(\ell)}}\leq  c \sum_{i} \vectorsym{h}^{(\layer)}_{i} \brackets{\lw_i - \quantlw_i}^2,
\end{equation}
where 
$\vectorsym{h}^{(\layer)}_{i}=\squareb{\expectation{\matsym{J}^{\brackets{\lw}}\brackets{\x}^T\matsym{J}^{\brackets{\lw}}\brackets{\x}}{\x}}_{i,i}$ is a vector representing the diagonal elements of $\expectation{\matsym{J}^{\brackets{\lw}}\brackets{\x}^T\matsym{J}^{\brackets{\lw}}\brackets{\x}}{\x}$.
\qed

%% file: files/appendix/hessian_apendix.tex
\section{Label-Free Hessian Details}

\subsection{Hessian matrix approximation}
\label{appx:hessian-process}

Following the definition of the Hessian matrix from Equation~\ref{eq:hessian-matrix} and using the chain rule as in \cite{li2021brecq} we obtain the $\brackets{i, j}$ element in $\hessian{\zf}$ by:
\begin{equation}
    \begin{alignedat}{2}
    \nonumber&\squareb{\hessian{\zf}\brackets{\x,\y}}_{i,j}
    =
    \frac{\partial^2\mathcal{L}_{task}\brackets{\y,f\brackets{\x}}}{\partial\idxz{i}\partial\idxz{j}} \\      
    &=
    \sum_{k=1}^{d_0}\squareb{g\brackets{\x, \y}}_k \cdot
    \frac{\partial^2\squareb{f\brackets{\x}}_k}{\partial\idxz{i}\partial\idxz{j}} +
    \frac{\partial f\brackets{\x}}{\partial\idxz{i} }^T \cdot
    \at{\frac{\partial^2\mathcal{L}_{task}\brackets{\y,\vectorsym{r} }}{\partial\vectorsym{r}\partial\vectorsym{r}^T}}{\vectorsym{r}=f\brackets{\x}} \cdot
    \frac{\partial f\brackets{\x}}{\partial\idxz{j}},
    \end{alignedat}
\end{equation}
 where $g\brackets{\x, \y} = \at{\nabla_{\vectorsym{r}}\mathcal{L}_{task}\brackets{\y,\vectorsym{r} }}{\vectorsym{r}=f\brackets{\x}}$
is the gradient of the task loss w.r.t. the network output.

Following the assumption that for a pre-trained network, the weights have converged to minimize the task loss, thus the gradients are close to~0~\cite{yao2021hawq, nagel2020up, li2021brecq},
we obtain the Hessian matrix approximation result presented in Equation~\ref{eq:hessian-approx}.

\subsection{Loss functions Hessian overview}
\label{apx:label_free_hessian}

In Section \ref{sec:sample-layer-attn} we claim that for many loss functions, the Hessian matrix of the task loss w.r.t. the model output is dependent exclusively on the model output. 
This implies that we can compute the Hessian without needing to compute the derivative of the original loss function. Here, we prove this claim by presenting a closed-form equation of the Hessian matrix for a set of commonly used loss functions. 

It can be observed that for several common loss functions such as Mean-Squared Error (MSE), Cross Entropy with Softmax, Binary Cross-Entropy Sigmoid, Poisson NLL\footnote{We use a version of Poisson NNL with a log input. For more details see \url{https://pytorch.org/docs/stable/generated/torch.nn.PoissonNLLLoss.html}} and Gaussian NLL the Hessian is only dependent on the model output. 
Table~\ref{tab:hessian} shows the corresponding Hessian for each of the aforementioned functions.

\begin{table}[H]
\centering
\caption{Loss function with their corresponding Hessian}
\label{tab:hessian}
\begin{tabular}{|c|c|c|}
\hline
\textbf{Loss}        & \textbf{Hessian}   & \textbf{Upper Bound}                                        \\ \hline
MSE         & $\matsym{A}\brackets{\vectorsym{r}}=\frac{2}{d_0}\matsym{I}_{d_0}$ & $\frac{2}{d_0}\matsym{I}$\ \\ \hline
CE-Softmax &  $\squareb{\matsym{A}\brackets{\vectorsym{r}}}_{i,j}=\begin{cases}\squareb{\mathrm{SM}\brackets{\vectorsym{r}}}_i\cdot\squareb{\mathrm{SM}\brackets{\vectorsym{r}}}_j,  &i\neq j\\\squareb{\mathrm{SM}\brackets{\vectorsym{r}}}_i\cdot\brackets{1-\squareb{\mathrm{SM}\brackets{\vectorsym{r}}}_i}, &i=j\end{cases}$ & $\matsym{I}$ \\ \hline
BCE-Sigmoid & $\matsym{A}\brackets{\vectorsym{r}}=\mathrm{Diag}\brackets{\mathrm{Sigmoid}\brackets{\vectorsym{r}}\odot\brackets{1-\mathrm{Sigmoid}\brackets{\vectorsym{r}}}}$   & $\matsym{I}$                                               \\ \hline
GaussianNLL &                                                 $\matsym{A}\brackets{\vectorsym{r}}=\frac{2}{\sigma^2}\matsym{I}_n$ & $\frac{2}{\sigma^2}\matsym{I}$  \\ \hline
PoissonNLL  & $\matsym{A}\brackets{\vectorsym{r}}=\mathrm{Diag}\brackets{\exp\brackets{\vectorsym{r}}}$                                                 & - \\ \hline
\end{tabular}
\end{table}
where $\odot$ is the Hadamard product, $\squareb{\mathrm{SM}\brackets{\x}}_k=\frac{\exp\brackets{\squareb{\x}_k}}{\sum_{j=1}^n\exp\brackets{\squareb{\x}_j}}$ is the softmax function, $\mathrm{Sigmoid}\brackets{x}$ is the sigmoid function and $\sigma^2$ is the varinace of the Gaussian NLL loss. In Table ~\ref{tab:hessian}, $\mathrm{Diag}\brackets{\vectorsym{r}}$ denote the transform of vector into a diagonal matrix~$\matsym{D}$ which is defined as: 
$$\squareb{\matsym{D}}_{i,j}=\begin{cases}0, & i\neq j\\
                                        \vectorsym{r}_i, & i=j\end{cases}$$

We provide a detailed proof of the Hessian computation of the loss functions presented in Table~\ref{tab:hessian}.
\begin{proof}
Denote the label vector as $\y\in\mathbb{R}^{d_0}$ and the prediction vector as $\vectorsym{r}\in\mathbb{R}^{d_0}$. 

\begin{itemize}
    \item \textit{Mean-Squared Error}: $\mathcal{L}\brackets{\y,\vectorsym{r}}=\frac{1}{d_0}\sum_{i=
    1}^{d_0}\brackets{\vectorsym{r}_i-\vectorsym{y}_i}^2$ and its Hessian is given by: 
        \begin{align}
            \squareb{\matsym{B}\brackets{\y,\vectorsym{r}}}_{k,j}&=\frac{\partial^2\mathcal{L}\brackets{\y,\vectorsym{r}}}{\partial\vectorsym{r}_k\partial\vectorsym{r}_j}=\frac{\partial}{\partial\vectorsym{r}_k}\frac{2}{d_0}\brackets{\vectorsym{r}_j-\vectorsym{y}_j}\nonumber\\
            &=
                \begin{cases}
                         0,   &  k\neq j\\
                          \frac{2}{d_0},  &  k=j
                \end{cases}
            \xrightarrow{}\matsym{B}\brackets{\y,\vectorsym{r}}=\frac{2}{d_0}\matsym{I}
        \end{align}

    \item \textit{Cross-Entropy with Softmax}: 
    \begin{equation*}
        \mathcal{L}\brackets{\y,\vectorsym{r}}=-\sum_{i=1}^{d_0}\brackets{\vectorsym{y}_i\log{\brackets{\squareb{\mathrm{SM}\brackets{\vectorsym{r}}}_i}}}=-\sum_{i=
    1}^{d_0}\vectorsym{y}_i\cdot \vectorsym{r}_i+\vectorsym{y}_i\log\brackets{\sum_{l=1}^{d_0}\exp\brackets{\vectorsym{r}_l}}.
    \end{equation*}
    Note that since $\vectorsym{y}$ is a distribution vector it holds that $\sum_{i=1}^{d_0}\vectorsym{y}_i = 1$.
    Then, the Hessian is given by: 
        \begin{align}
            \begin{alignedat}{1}
                &\squareb{\matsym{B}\brackets{\y,\vectorsym{r}}}_{k,j}= 
                \frac{\partial^2\mathcal{L}\brackets{\y,\vectorsym{r}}}{\partial\vectorsym{r}_k\partial\vectorsym{r}_j} =
                \frac{\partial}{\partial\vectorsym{r}_k}\brackets{-\vectorsym{y}_j+\sum_{i=1}^{d_0}\frac{\exp\brackets{\vectorsym{r}_j}}{\sum_{l=1}^{d_0}\exp\brackets{\vectorsym{r}_l}}\vectorsym{y}_i}\\
                &=\frac{\partial}{\partial\vectorsym{r}_k}\frac{\exp\brackets{\vectorsym{r}_j}}{\sum_{l=1}^{d_0}\exp\brackets{\vectorsym{r}_l}} =\begin{cases}
                    -\squareb{\mathrm{SM}\brackets{\vectorsym{r}}}_k\cdot\squareb{\mathrm{SM}\brackets{\vectorsym{r}}}_j,  &k\neq j\\\squareb{\mathrm{SM}\brackets{\vectorsym{r}}}_k\cdot\brackets{1-\squareb{\mathrm{SM}\brackets{\vectorsym{r}}}_k}, &k=j
                \end{cases}
            \end{alignedat}
        \end{align}

    \item \textit{Binary Cross-Entropy with Sigmoid}: 
    \begin{equation*}
        \mathcal{L}\brackets{\y,\vectorsym{r}}=-\sum_{i=1}^{d_0}\brackets{\vectorsym{y}_i\log{\brackets{\mathrm{Sigmoid}\brackets{\vectorsym{r}_i}}}}=\sum_{i=1}^{d_0}\brackets{\vectorsym{y}_i\cdot\log{\brackets{1+\exp{\brackets{-\vectorsym{r}_i}}}}},
    \end{equation*}
     and its Hessian is given by: 
        \begin{align}
            \begin{alignedat}{1}\squareb{\matsym{B}\brackets{\y,\vectorsym{r}}}_{k,j}&=\frac{\partial^2\mathcal{L}\brackets{\y,\vectorsym{r}}}{\partial\vectorsym{r}_k\partial\vectorsym{r}_j} =-\frac{\partial}{\partial\vectorsym{r}_k}\brackets{\vectorsym{y}_i\cdot\frac{\exp\brackets{-\vectorsym{r}_i}}{1+\exp\brackets{-\vectorsym{r}_i}}} \\ &=
                \begin{cases}
                    0, & k\neq j\\\mathrm{Sigmoid}\brackets{\vectorsym{r}_k}\cdot\brackets{1-\mathrm{Sigmoid}\brackets{\vectorsym{r}_k}}, & k=j
                \end{cases} \\
            &\xrightarrow{}\matsym{B}\brackets{\y,\vectorsym{r}}=\mathrm{Diag}\brackets{\mathrm{Sigmoid}\brackets{\vectorsym{r}}\odot\brackets{1-\mathrm{Sigmoid}\brackets{\vectorsym{r}}}}
            \end{alignedat}
        \end{align}
    \item\textit{Gaussian Negative Log Likelihood}: $\mathcal{L}\brackets{\y,\vectorsym{r}}=\frac{1}{\sigma^2}\sum_{i=
    1}^{d_0}\brackets{\vectorsym{r}_i-\vectorsym{y}_i}^2$ and its Hessian is given by: 
    \begin{align}
        \squareb{\matsym{B}\brackets{\y,\vectorsym{r}}}_{k,j}&=\frac{\partial^2\mathcal{L}\brackets{\y,\vectorsym{r}}}{\partial\vectorsym{r}_k\partial\vectorsym{r}_j}=\frac{\partial}{\partial\vectorsym{r}_k}\frac{2}{\sigma^2}\brackets{\vectorsym{r}_j-\vectorsym{y}_j}\nonumber\\
        &=\begin{cases}
                 0, & k\neq j\\
                  \frac{2}{\sigma^2}, & k=j
        \end{cases}\xrightarrow{}\matsym{B}\brackets{\y,\vectorsym{r}}=\frac{2}{\sigma^2}\matsym{I}
    \end{align}

\item\textit{Poisson Negative Log Likelihood}: $\mathcal{L}\brackets{\y,\vectorsym{r}}=\sum_{i=
    1}^{d_0}\exp\brackets{\vectorsym{r}_i}-\vectorsym{y}_i\cdot \vectorsym{r}_i$ and its Hessian is given by: 

    \begin{align}
        \begin{alignedat}{1}
        \squareb{\matsym{B}\brackets{\y,\vectorsym{r}}}_{k,j}&=\frac{\partial^2\mathcal{L}\brackets{\y,\vectorsym{r}}}{\partial\vectorsym{r}_k\partial\vectorsym{r}_j} =\frac{\partial}{\partial\vectorsym{r}_k}\brackets{\exp\brackets{\vectorsym{r}_j}-\vectorsym{y}_j}\\&=
            \begin{cases}
                0, & k\neq j \\
                \exp\brackets{\vectorsym{r}_i}, & k=j
            \end{cases}
        \xrightarrow{}\matsym{B}\brackets{\y,\vectorsym{r}}=\mathrm{Diag}\brackets{\exp\brackets{\vectorsym{r}}}
        \end{alignedat}
    \end{align}

\end{itemize}

\end{proof}

\subsection{Hessian approximation comparison}
\begin{figure}[H]
    \begin{subfigure}[t]{0.48\textwidth}
        \centering
        \includegraphics[width=\textwidth]{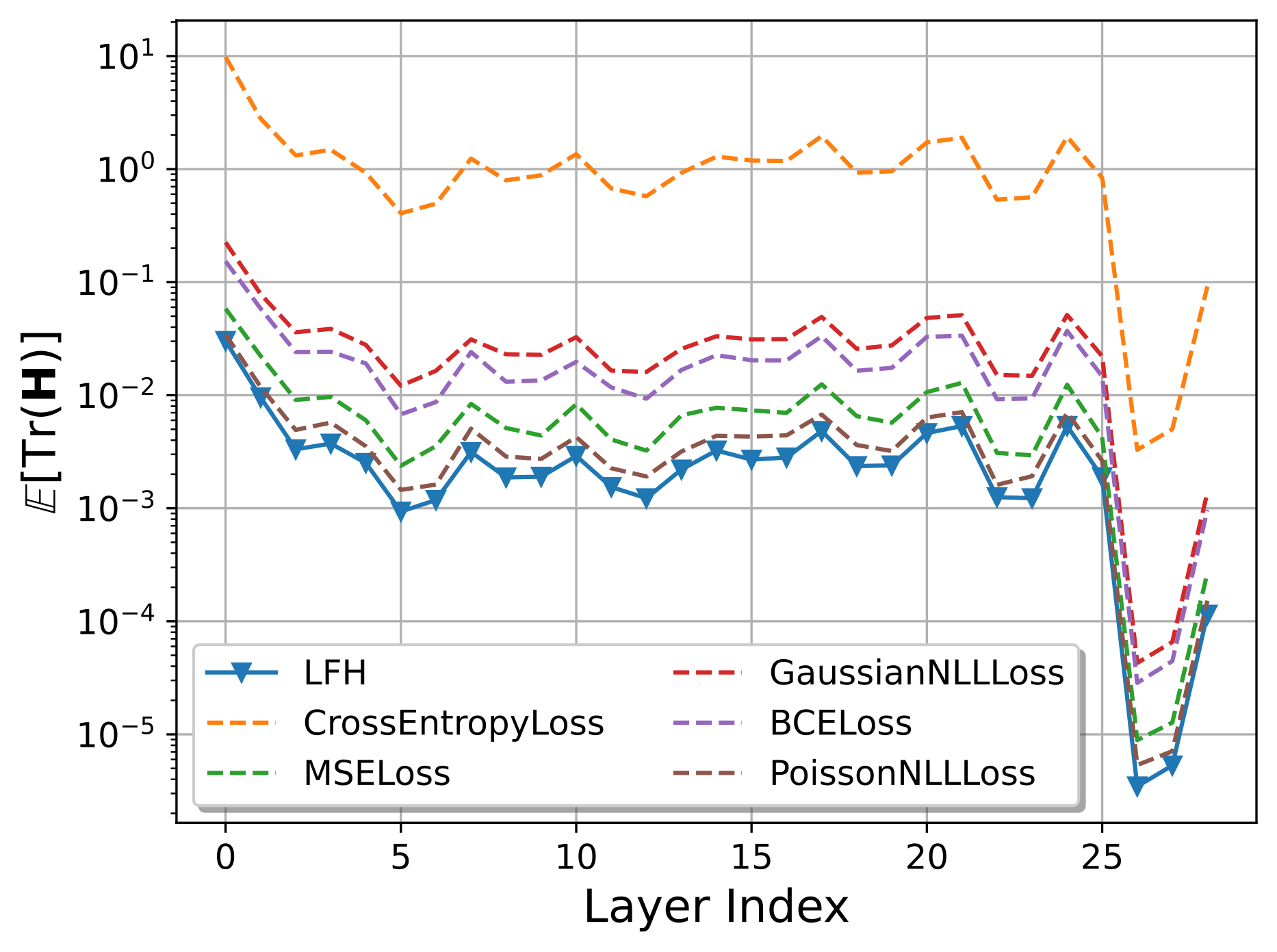}
        \caption{Unnormalized}
    \label{fig:resnet-hessian-approx}
    \end{subfigure}
    \hfill
    \begin{subfigure}[t]{0.48\textwidth}
        \centering
        \includegraphics[width=\textwidth]{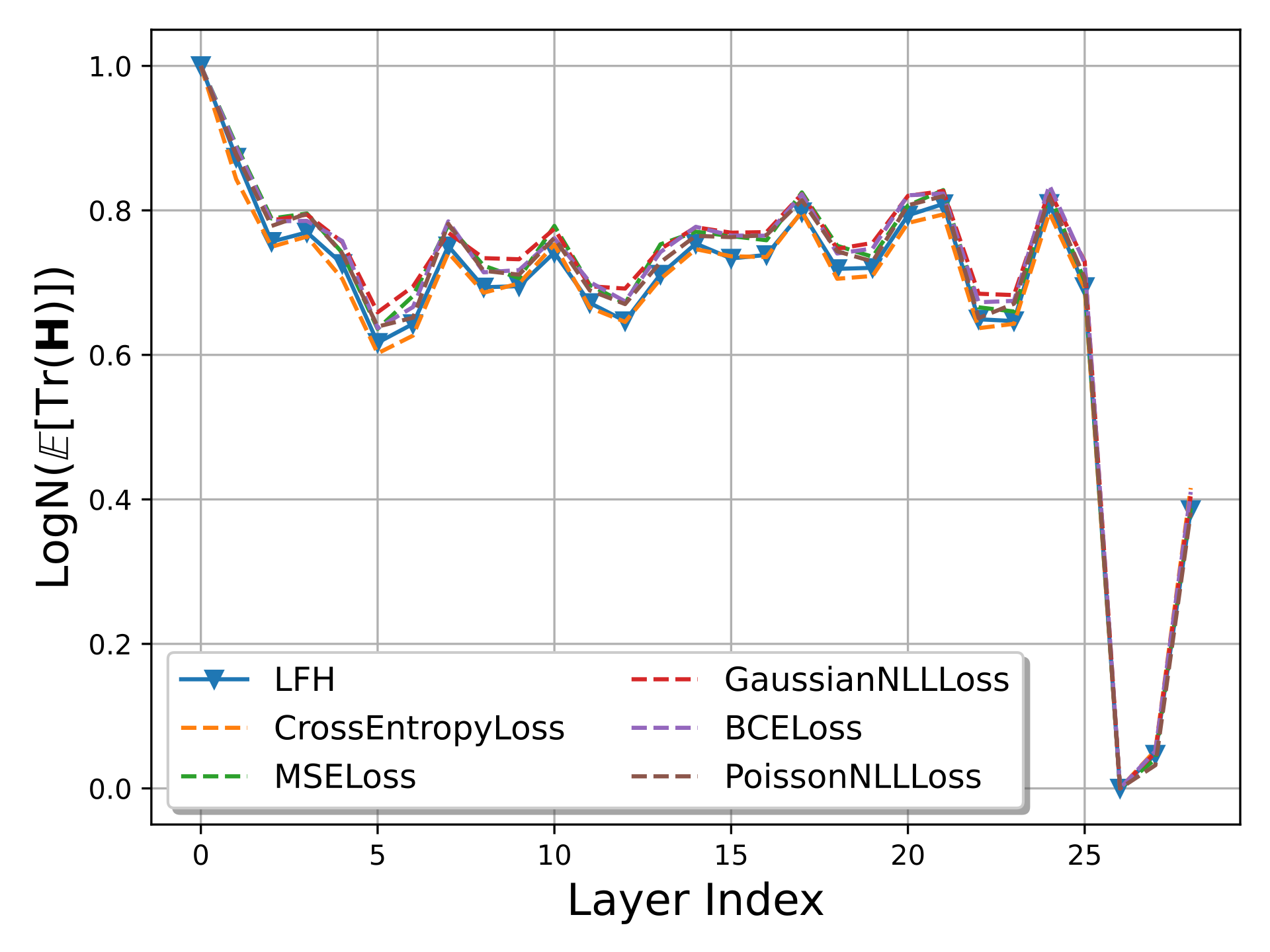}
        \caption{Normalized}
        \label{fig:hessian-logscale}
    \end{subfigure}
    \caption{Comparison between Hessian trace approximation following the LFH assumption, and the true Hessian trace~\cite{yao2020pyhessian} of different loss functions on ResNet18 on random ImageNet samples.
    In (a) the values of the Hessian trace are normalized, while in (b) the values are normalized using log normalization.)}
    \label{fig:different_loss}
\end{figure}

We demonstrate the Label-Free Hessian approximation compared to the actual Hessian values for different loss functions on ResNet18 in Figure~\ref{fig:different_loss}.
For simplicity of exposition, we present the averaged Hessian trace evaluation taken over a set of 256 random samples from ImageNet.

In (a), we can observe that there is a constant scaling gap between the LFH trace value and the true Hessian trace.
This gap can be bridged by applying vector-wise log normalization as demonstrated in (b), 
i.e., $LogN(\vectorsym{v}) = \frac{ln\brackets{\vectorsym{v}} - c_{min}}{c_{max} - {c_{min}}}$,
where $\vectorsym{v}$ is a vector, $ln$ is an element-wise logarithmic function, and $c_{max}$ and $c_{min}$ are the logarithmic values of $\max_i{\vectorsym{v}_i}$ and $\min_i{\vectorsym{v}_i}$
respectively. 
The log normalization corrects the loss shift
and closes the gap between the approximation and the actual
Hessian trace.

\begin{figure}[H]
 \centering
    \begin{subfigure}{0.48\textwidth}
        \centering

        \includegraphics[width=1.0\textwidth]{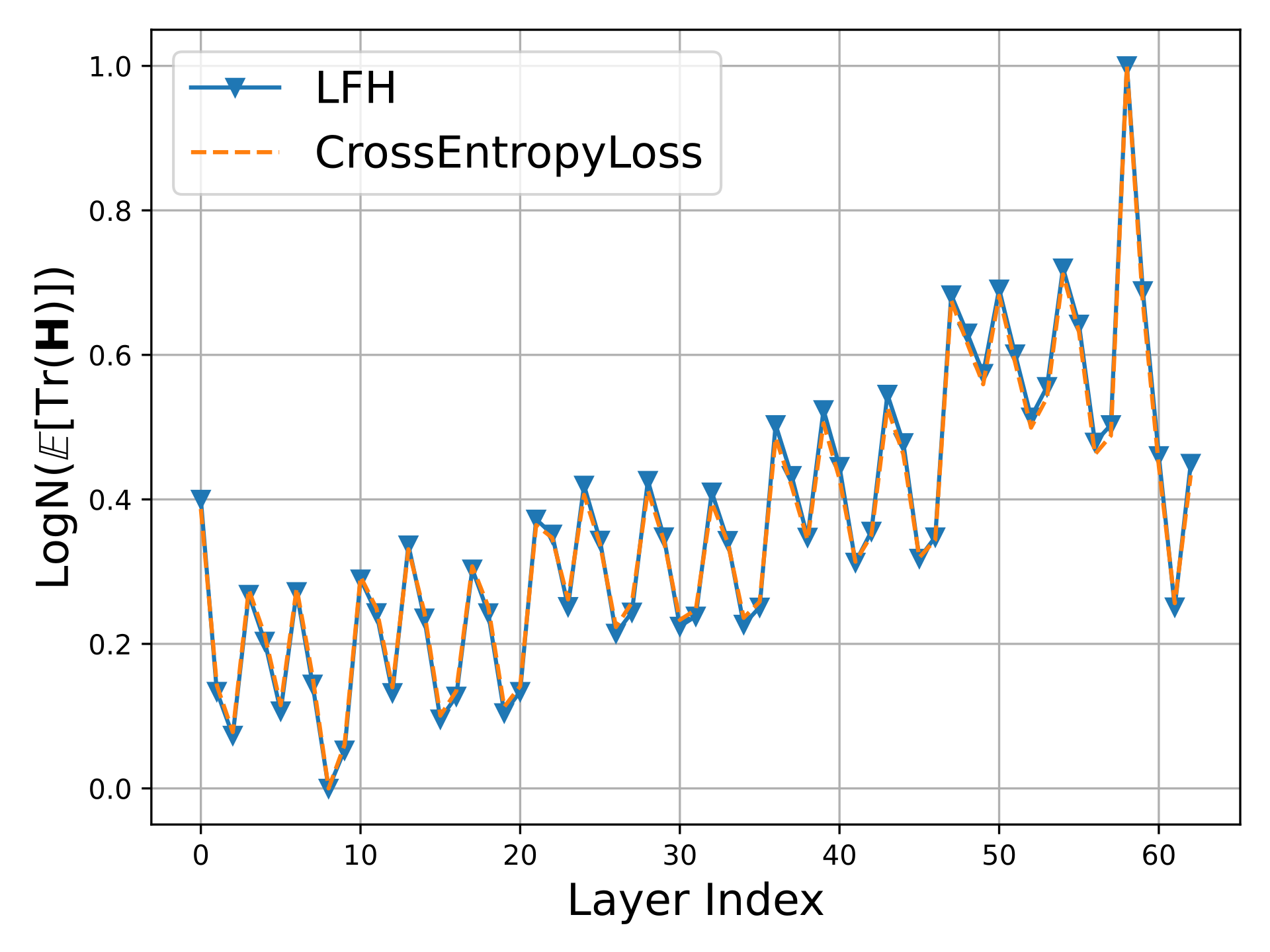}
        \caption{MobileNet-V2}
    \end{subfigure}
    \begin{subfigure}{0.48\textwidth}
        \centering
        \includegraphics[width=1.0\textwidth]{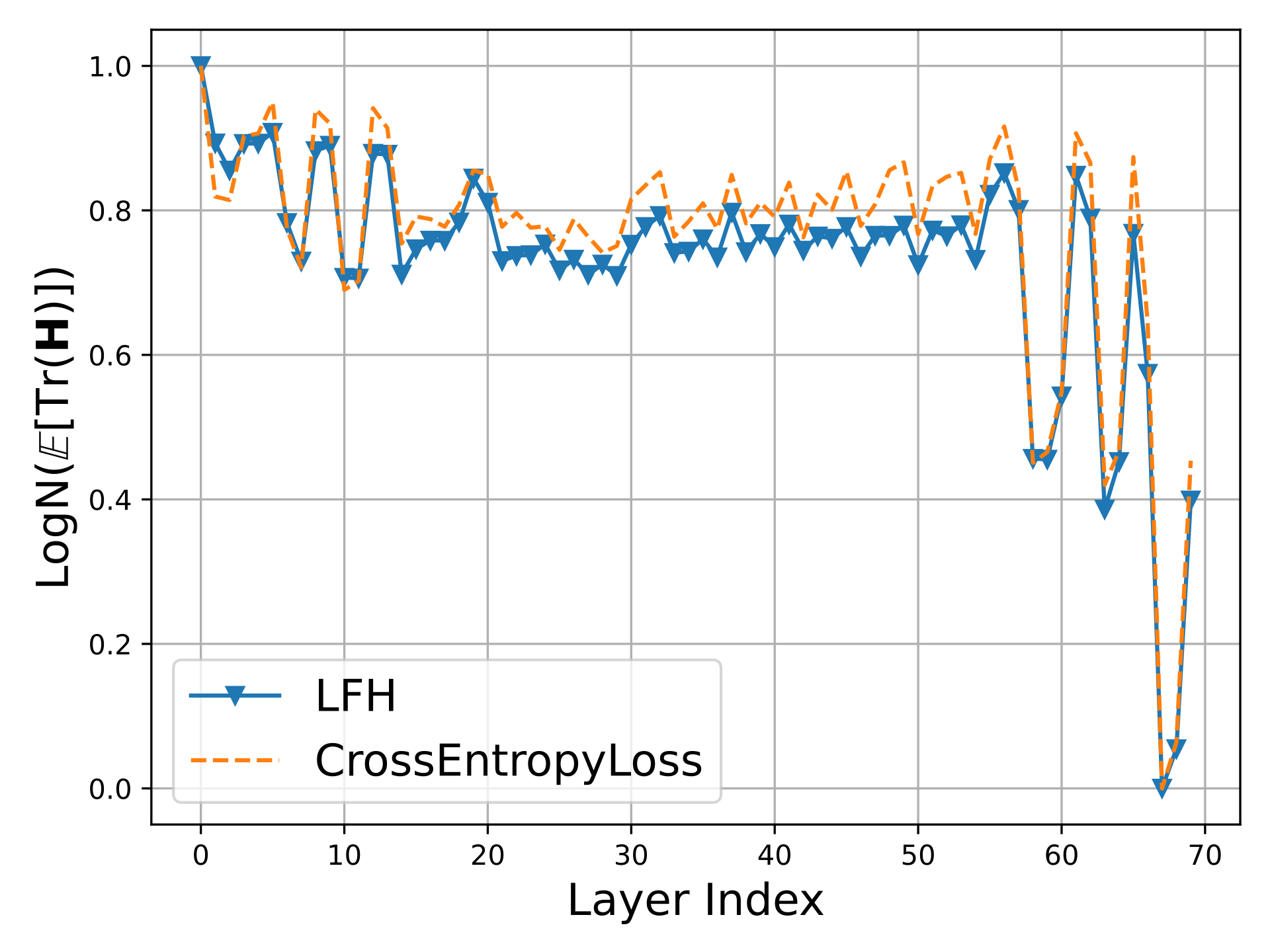}
        \caption{ResNet50}
    \end{subfigure}
    \caption{A comparison between the Hessian trace of common task loss functions and our task-invariant \emph{Label-Free Hessian} approximation on MobileNet-V2 and ResNet50, computed on a random set from ImageNet.}
    \label{fig:more-hessian-apprx}
\end{figure}

A similar comparison against the Cross-Entropy loss function, for MobileNetV2 and ResNet50, is presented in Figure~\ref{fig:more-hessian-apprx}.

\label{appx:hessian-approx-figs}

\subsection{Efficient computation of Hessian diagonal and max value}
\label{apx:efficient_jac}
Here, we suggest using the Hutchinson algorithm \cite{avron2011randomized}, which efficiently computes an approximation of the diagonal of a symmetric positive semi-definite matrix. 
We aim to compute the Hessian matrix bound in Equation~\ref{eq:lfh}. 
Computing the Jacobian matrix of the model w.r.t. an activation tensor can be expensive, specifically when performed for all activation tensors on which we measure the difference from the floating point model.
We avoid the necessity of computing the exact Jacobian matrix by utilizing the following known \cite{yao2020pyhessian} mathematical relation:
\begin{equation}
    \vectorsym{h}_v^{(\zf)}\brackets{\vectorsym{v},\vectorsym{x}}=\divc{\brackets{\vectorsym{v}^T f\brackets{\vectorsym{x}}}}{\zf}=
    \divc{\vectorsym{v}^T}{\zf}f\brackets{\vectorsym{x}}+\vectorsym{v}^T\divc{f\brackets{\vectorsym{x}}}{\zf}=
    \vectorsym{v}^T\matsym{J}^{(\zf)}\brackets{\vectorsym{x}},
\end{equation}
where $\vectorsym{v}\in\mathbb{R}^{d_0}$ is a random vector with a known distribution. 
If we choose the distribution of $\vectorsym{v}$ such that~$\expectation{\vectorsym{v}\vectorsym{v}^T}{\vectorsym{v}} = {\matsym{I}_{d_0}}$, where $\matsym{I}_{d_0}$ is an identity matrix of size $d_0$, then:
\begin{align}
\label{eq:norm2}
    \matsym{J}^{\brackets{\zf}}\brackets{\x}^T\matsym{J}^{\brackets{\zf}}\brackets{\x}
    &=\matsym{J}^{\brackets{\zf}}\brackets{\x}^T\matsym{I}_{d_0}\matsym{J}^{\brackets{\zf}}\brackets{\x}\\\nonumber&=\matsym{J}^{\brackets{\zf}}\brackets{\x}^T\expectation{\vectorsym{v}\vectorsym{v}^T}{\vectorsym{v}}\matsym{J}^{\brackets{\zf}}\brackets{\x}\\\nonumber
    &=\expectation{\matsym{J}^{\brackets{\zf}}\brackets{\x}^T\vectorsym{v}\vectorsym{v}^T\matsym{J}^{\brackets{\zf}}\brackets{\x}}{\vectorsym{v}}\\\nonumber&=\expectation{\brackets{\vectorsym{h}_v^{(\zf)}\brackets{\vectorsym{v},\vectorsym{x}}}^T\vectorsym{h}_v^{(\zf)}\brackets{\vectorsym{v},\vectorsym{x}}}{\vectorsym{v}},
\end{align}
Using~\eqref{eq:norm2}, we approximate the expected Jacobian norm by using the empirical mean of $M$ different $\vectorsym{v}_m$ vectors, computed with a set of random vectors~${\vectorsym{v}_m\sim\normaldis{0}{\matsym{I}_m}}$. 
Then, the approximation is given by:
\begin{equation}\label{eq:approx_norm}
\expectation{\matsym{J}^{\brackets{\zf}}\brackets{\x}^T\vectorsym{v}\vectorsym{v}^T\matsym{J}^{\brackets{\zf}}\brackets{\x}}{\vectorsym{v}}\approx\frac{1}{M}\sum_{m=1}^{M}\brackets{\vectorsym{h}_{v_m}^{(\zf)}\brackets{\vectorsym{v},\vectorsym{x}}}^T \vectorsym{h}_{v_m}^{(\zf)}\brackets{\vectorsym{v},\vectorsym{x}}.
\end{equation}

\subsubsection{Diagonal approximation For {\hmse}}
In Algorithm \ref{alg:diag-lfh} we describe the computation of the bound in Equation~\ref{eq:hmse-loss} using the approximation from Equation~\ref{eq:approx_norm}.
The algorithm reduces the number of Jacobian matrices needed to be computed from $|\mathcal{D}_R|\cdot \numipts\cdot d_0$ to $|\mathcal{D}_R|\cdot \numipts\cdot M$. 
Since this algorithm converges in~${M\sim 50}$ iterations for most cases, and that $d_0>>M$ it means that for most cases the computation is much more tractable. 

\begin{algorithm}[H]
   \caption{Label-Free Hessian Diagonal Computation}
   \label{alg:diag-lfh}
    \begin{algorithmic}
       \STATE {\bfseries Input:}  Batch of samples $\mathcal{D}_{R}$, model $f$, weights tensor $\lw$
       \STATE  $\vectorsym{h}^{(\layer)} \gets 0$.
       \FOR{$\vectorsym{x}\in\mathcal{D}_R$}
       \STATE $\vectorsym{y}=f\brackets{\vectorsym{x}}$
        \FOR{all $k=1,2,...,M$}
            \STATE $\vectorsym{v}\sim\mathcal{N}\brackets{0,\matsym{I}_m}$
            \STATE  $\y_{\vectorsym{v}_m}=\vectorsym{v}^T\vectorsym{y}$ 
            \STATE $\vectorsym{h}_{v_m}^{(\lw)}=\divc{\y_{\vectorsym{v}_m}}{\lw}$
            \STATE $\vectorsym{h}^{(\layer)}\gets \vectorsym{h}^{(\layer)} + \frac{1}{M \abs{\mathcal{D}_R}}\mathrm{Diag}\brackets{\brackets{\vectorsym{h}_{v_m}^{(\lw)}}^T\vectorsym{h}_{v_m}^{(\lw)}}$ using \eqref{eq:approx_norm}
        \ENDFOR
       \ENDFOR
    \end{algorithmic}
\end{algorithm}

\subsubsection{Maximal Diagonal Element For SLA}
In Algorithm \ref{alg:max-lfh} we describe the computation of the bound in Equation~\ref{eq:sample-layer-attn} using the approximation in Equation~\ref{eq:approx_norm}.
Similarly to Algorithm~\ref{alg:diag-lfh}, the number of Jacobians needed to be computed is reduced to $|\mathcal{D}_R|\cdot \numipts\cdot M$, making the computation much more tractable.
Nevertheless, for the SLA bound we compute a score per-sample, therefore, we perform a batch computation of the Jacobians approximation.

\begin{algorithm}[H]
   \caption{Label-Free Max Diagonal Value Computation}
   \label{alg:max-lfh}
    \begin{algorithmic}
       \STATE {\bfseries Input:}  Batch of samples $\mathcal{D}_{R}$, model $f$, activation tensor $\lz$
       
       \FOR{$\vectorsym{x}\in\mathcal{D}_R$}
       \STATE $\vectorsym{y}=f\brackets{\vectorsym{x}}$
       \STATE  $\vectorsym{h}^{(\layer)} \gets 0$.
        \FOR{all $m=1,2,...,M$}
            \STATE $\vectorsym{v}_m\sim\mathcal{N}\brackets{0,\matsym{I}_m}$
            \STATE  $\y_{\vectorsym{v}_m}=\vectorsym{v}^T\vectorsym{y}$ 
            \STATE $\vectorsym{h}_{v_m}^{(\lz)}=\divc{\y_{\vectorsym{v}_m}}{\lz}$
            \STATE $\vectorsym{h}^{(\layer)}\gets \vectorsym{h}^{(\layer)} + \frac{1}{M}\mathrm{Diag}\brackets{\brackets{\vectorsym{h}_{v_m}^{(\lw)}}^T\vectorsym{h}_{v_m}^{(\lw)}}$ using \eqref{eq:approx_norm}
        \ENDFOR
         \STATE $\maxhess{\layer}\brackets{{\vectorsym{x}}}\gets \max\limits_{i}\vectorsym{h}^{(\layer)}_i$
       \ENDFOR
    \end{algorithmic}
\end{algorithm}

%% file: files/appendix/additional_results.tex
\section{Additional Experimental Results}
\label{apx:add-res}

In Table~\ref{tab:res-more-samples} we present an additional comparison of our method {\name} against a recently published work by Ma et al.~\cite{ma2023solving}.
We split these results from the main comparison presented in Table~\ref{tab:all-res} for the sake of a fair comparison since this work uses a larger representative dataset for the optimization.

\begin{table}[H]
\centering
\caption{Comparison between {\name} and~\cite{ma2023solving}, which uses an increased representative dataset of 4,096 samples for ResNet18 and MobileNetV2 and 2048 samples for ResNet50, on ImageNet classification task.}
\label{tab:res-more-samples}
\resizebox{0.8\textwidth}{!}{%
\begin{tabular}{lcccc}
\hline
\textbf{Methods} & \textbf{Bits (W/A)} & \textbf{ResNet18}                                             & \textbf{ResNet50}                                             & \textbf{MobileNetV2}                                          \\ \hline
Full Prec.       & 32/32               & 71.01                                                         & 76.63                                                         & 72.62                                                         \\ \hline
Ma et al.        & \multirow{2}{*}{4/4}                 & 69.46                                                         & 75.35                                                         & 68.84                                                         \\
\textbf{\our}     &                  & \textbf{69.71} \tstd{0.07} &  \textbf{75.46} \tstd{0.09} & \textbf{70.01} \tstd{0.05} \\ \hline
Ma et al.        & \multirow{2}{*}{3/3}                 & 66.30                                                         & 71.92                                                         & 58.40                                                         \\
\textbf{\our}     &                 & \textbf{67.11} \tstd{0.11}                                                & \textbf{72.71} \tstd{0.09}                                                & \textbf{62.69} \tstd{0.22}                                               \\ \hline
\end{tabular}
}
\end{table}